\documentclass[11pt]{article}

\input{packages}

\usepackage[ruled]{algorithm2e}

\usepackage{mathrsfs} 
\usepackage{soul} 
\newcommand{\inparen}[1]{\left(#1\right)}

\newcommand{\tv}[2]{\operatorname{d}_{\mathsf{TV}}\inparen{#1,#2}}

\newcommand{\supp}{\operatorname{supp}}
\newcommand{\iid}{{\rm i.i.d.}}
\newcommand{\unif}{{\rm Unif}}
\newcommand{\law}{\mathrm{Law}}

\newcommand{\bin}{\mathrm{Bin}}

\newcommand{\unifcube}{\unif(\{\pm 1\}^n)}
\newcommand{\lsb}{\mathrm{LSB}}
\newcommand{\good}{\mathrm{good}}
\newcommand{\negl}{\mathsf{negl}}

\newcommand{\onl}{\text{onl}}
\newcommand{\cnt}{\text{count}}
\newcommand{\gen}{\text{gen}}
\newcommand{\sq}{\text{SQ}}
\newcommand{\erm}{\text{ERM}}
\newcommand{\pac}{\text{PAC}}
\newcommand{\den}{\text{den}}

\title{PAC Learning to Sample:\\
On Computational Aspects of Language Generation}
 
\author{
\begin{tabular}{ccc}
Alkis Kalavasis\footnote{Work done when AK was supported by the FDS Postdoctoral Fellowship at Yale University.}
&
Shuchen Li
&
Manolis Zampetakis
\\
Harvard University
&
Yale University
&
Yale University
\\
\small \texttt{\textcolor{magenta}{alvertos\_kalavasis@seas.harvard.edu}}
&
\small \texttt{\textcolor{magenta}{shuchen.li@yale.edu}}
&
\small \texttt{\textcolor{magenta}{manolis.zampetakis@yale.edu}}
\end{tabular}
}

\begin{document}

\maketitle

\begin{abstract}
The success of modern language models has motivated a growing theoretical literature on learning to sample from distributions on discrete domains. A prominent and widely adopted abstraction is the model of language generation in the limit, introduced by \citet{kleinberg2024language}, which studies eventually consistent generation: after some finite but unspecified training time on positive examples, the algorithm must generate from the true target language. While this model is theoretically elegant and the positive results are surprisingly general, contrasting classical works of \citet{gold1967language,angluin1980inductive} on language identification in the limit, most guarantees are of \emph{qualitative} nature.
In this work, we shift towards a different model of language generation, inspired by statistical and computational learning theory, aiming for \emph{quantitative} results in terms of sample and time complexity. 

To this end, we study a PAC model of language generation, proposed (in a different context) by \citet{de2014learning}, in which a Boolean concept class $C$ over $\{-1,1\}^n$ defines the target languages and the problem is defined as follows: Given uniformly random positive examples of an unknown concept $f \in C$, the learner must output a sampler whose output distribution (conditioned on the training data) is close in total variation distance to the uniform distribution over $f^{-1}(1)$, i.e., the set of satisfying assignments of $f$. Prior work implicitly gives an efficient algorithm for the task of PAC learning to sample $C$ assuming access to three algorithmic primitives: an online learner for $C$, approximate sampling and counting oracles for $C$, and an SQ learner for $C$. The algorithm that learns to sample from $C$ runs in time polynomial in the time required to implement these oracles. 

We ask whether these ingredients, and in particular online learnability, are necessary for efficiently PAC learning to sample. First, we show that there exists a Boolean concept class $C$ for which online learning is cryptographically hard, yet PAC learning to sample is computationally efficient. Motivated by this separation, we develop new algorithms for PAC learning to sample that use offline ERM oracles (instead of online oracles), together with approximate sampling/counting oracles for intersections of concepts in $C$ and an SQ learner for $C$. We view our results, {together with the work of \citet{de2014learning}}, as a first step towards understanding the computational complexity of language generation. 
\end{abstract}

   \newpage 
   
\section{Introduction}

Modern language models, after being trained on an extremely large corpus of text, learn to generate content that resembles human language, a capability that underlies a wide range of applications including dialogue systems, code generation, and  mathematical reasoning.
Their striking empirical success  therefore raises a fundamental theoretical question: what properties of a data distribution make \emph{efficient learning to sample} possible?

The success of this paradigm has motivated an extensive line of works that aim at
providing a theoretical understanding on {learning to sample}. One of the most useful and adopted theoretical abstractions is the model of language generation in the limit, introduced by \citet{kleinberg2024language}, which studies \emph{eventually} consistent generation: given a class of languages $\cL$, the learner observes (positive) examples from an unknown true language $K \in \cL$ and the goal is to design an algorithm that is guaranteed to generate new unseen examples from the true target language after a finite -yet unspecified- training time. In this regard, the algorithmic results in this model are mainly of qualitative nature.

Language generation in the limit builds upon classical works on inductive inference \cite{solomonoff1964formal}, initiated by  \citet{gold1967language} and further extended by \citet{angluin1980inductive}. These learning frameworks rely on three principles \cite{wigderson2019mathematics}: the data (i.e., the positive examples) arrive adversarially, the training time is finite but unspecified, and the learner has to eventually stop making mistakes. Soon after these seminal works, statistical approaches, initiated by Vapnik and Chervonenkis \cite{vapnik2013nature} and \citet{valiant1984theory} changed the focus and proposed three different guiding principles: the data are generated randomly, the amount of training samples and computation should be specific (and reasonable), and the learner is allowed to make mistakes at inference as long as they occur with sufficiently small probability. This led to the rich theory of Probably Approximately Correct (PAC) learning that allowed the community to transfer
the qualitative results on learnability (e.g., \cite{gold1967language,angluin1980inductive}) to more quantitative bounds, explicitly controlling the sample and time complexity of learning tasks \cite{valiant1984theory,vapnik2013nature,blumer1989learnability,devroye2013probabilistic,servedio2025probably,kearns1994introduction,shalev2014understanding}.

Before zooming into the question of the PAC analogue for language generation in the limit, it is worth mentioning that 
generative modeling has a clear statistical  formulation that appears already in the early works of learning theory by Kearns, Mansour, Ron, Rubinfeld, Schapire, and Sellie \cite{kearns1994learnability}: Given $n$ i.i.d.\ samples from an unknown data distribution $P$ on some domain $X$, output a sampler (or generator) $G_n : \{0,1\}^m \to X$
(i.e.,  a circuit  that gets as input some random seed (of length $m$) drawn from the uniform distribution $U$ and outputs samples $x \sim G_n(U)$ in $X$)
such that $d(P, G_n(U))$ is small with high probability, where $d(\cdot, \cdot)$ is some notion of distributional divergence. The main question is then: 
\begin{center}
    \emph{How does the sample and time complexity of learning to sample scale with the data distribution $P$?}
\end{center}

This question is very general and admits multiple answers when one focuses on a particular domain of application (e.g., image generation and diffusion models). Our goal in this work is to study, in particular, the PAC analogue of the work of \cite{kleinberg2024language} in  language generation, aiming at understanding the complexity of learning to generate from a target language given positive examples when the algorithm is limited to a reasonable amount of samples and time, a question that is an important aspect of multiple works on language acquisition \cite{gold1978complexity,clark2012computational,van2008tractable}. Hence, we will focus on discrete domains and on distributions $P$ that are associated to  ``languages''.
To this end, we have to find a 
conceptual bridge between the target distribution $P$ and the notion of a \emph{language} in the context of \cite{kleinberg2024language}. This leads us to the model of \emph{learning from satisfying assignments}, studied in \citet{de2014learning}, that offers such a natural connection, which we describe below.

\section{The PAC Language Generation Model of \cite{de2014learning}}

As in classical PAC learning, we have to put some constraints on the target $P$ to provide some inductive bias to the learner towards achieving the goal of (efficient) learning to sample. In our work, we study the model of \citet{de2014learning} in order to understand the above question by assuming that the target distribution $P$ is a ``low-complexity'' probability measure on the Boolean hypercube $\{-1,1\}^n$. We now explain the setting and the learning problem.

\begin{figure}[!ht]
    \centering
    \begin{tikzpicture}[scale=1.1, >=stealth]
  \definecolor{posregion}{RGB}{234, 246, 234}
  \definecolor{pospoint}{RGB}{0, 135, 40}
  \definecolor{negpoint}{RGB}{230, 20, 20}

  \fill[posregion] (0,0) -- (4.2, 3.4) -- (8.2, 0) -- cycle;

  \draw[->, thick] (-0.3, 0) -- (9.4, 0) node[below=2pt] {\large $x_1$};
  \draw[->, thick] (0, -0.3) -- (0, 6.2) node[left=4pt] {\large $x_2$};

  \draw[thick] (-0.315, -0.255) -- (6.3, 5.1);

  \draw[thick] (2.8, 4.53) -- (8.47, -0.2265);

  \node[align=center] (h1_text) at (6.6, 5.7) {\small $H_1$ boundary\\[-1pt]\small $(\mathbf{a}_1^\top \mathbf{x} = b_1)$};
  \draw[thick, ->, shorten >=2pt] (5.8, 5.5) to [out=180, in=90] (5.2, 4.21);

  \node[align=center] (h2_text) at (8.0, 2.8) {\small $H_2$ boundary\\[-1pt]\small $(\mathbf{a}_2^\top \mathbf{x} = b_2)$};
  \draw[thick, ->, shorten >=2pt] (7.3, 2.4) to [out=-150, in=50] (6.7, 1.3);

  \foreach \p in {
    (4.2, 2.9),
    (3.7, 2.3), (4.2, 2.3), (4.7, 2.3),
    (3.0, 1.8), (3.5, 1.8), (4.2, 1.7), (4.7, 1.8),
    (2.5, 1.3), (3.1, 1.3), (3.9, 1.3), (4.6, 1.2), (5.1, 1.3), (5.7, 1.3),
    (2.0, 0.8), (2.8, 0.8), (3.5, 0.8), (4.3, 0.8), (5.0, 0.8), (6.0, 0.8),
    (1.4, 0.4), (2.3, 0.4), (3.0, 0.4), (3.6, 0.3), (4.3, 0.3), (5.5, 0.4), (6.6, 0.3)
  } {
    \fill[pospoint] \p circle (2.2pt);
  }

  \foreach \p in {
    (0.5, 1.1), (0.7, 2.3), (0.8, 3.8), (0.9, 5.0),
    (1.2, 1.4), (1.4, 2.6), (1.5, 3.9), (1.4, 5.3),
    (1.8, 1.9), (2.0, 3.1), (2.1, 4.4), (2.0, 5.8),
    (2.5, 2.3), (2.7, 3.6), (2.8, 4.8), (2.8, 6.0),
    (3.2, 2.8), (3.4, 4.2), (3.5, 5.5),
    (3.8, 3.5), (4.2, 4.9), (4.3, 6.0),
    (4.6, 4.3), (5.2, 5.8),
    (5.7, 3.8), (6.4, 3.0), (6.2, 2.0), (7.0, 3.8), (7.0, 2.4), (7.5, 2.0), (7.8, 3.4),
    (7.6, 1.1), (8.2, 1.1), (8.3, 0.5)
  } {
    \fill[negpoint] \p circle (2.2pt);
  }

  \begin{scope}[shift={(-0.3, -1.2)}]
    \fill[posregion] (-0.5,-0.15) rectangle (0.0, 0.15);
    \node[anchor=west, align=left] at (0.1, -0.02) {\footnotesize Positive region: $H_1 \cap H_2$\\[-2pt]\footnotesize (intersection of two halfspaces)};

    \fill[pospoint] (4.6, 0) circle (2.5pt);
    \node[anchor=west] at (4.7, 0) {\footnotesize Positive examples $(+)$};

    \fill[negpoint] (7.8, 0) circle (2.5pt);
    \node[anchor=west] at (7.9, 0) {\footnotesize Negative examples $(+)$};
  \end{scope}

\end{tikzpicture}
    \captionsetup{margin=0.5cm}
    \caption{The figure shows a language $L$ induced by a hypothesis $f$ that is an intersection of two halfspaces. The elements of the language $L$ are the green points, where $h(x) = 1$. The red points are the negative examples, which do not belong to $L$. We are interested in learning in total variation the uniform distribution over the green points.}
    \label{fig:placeholder1}
\end{figure}
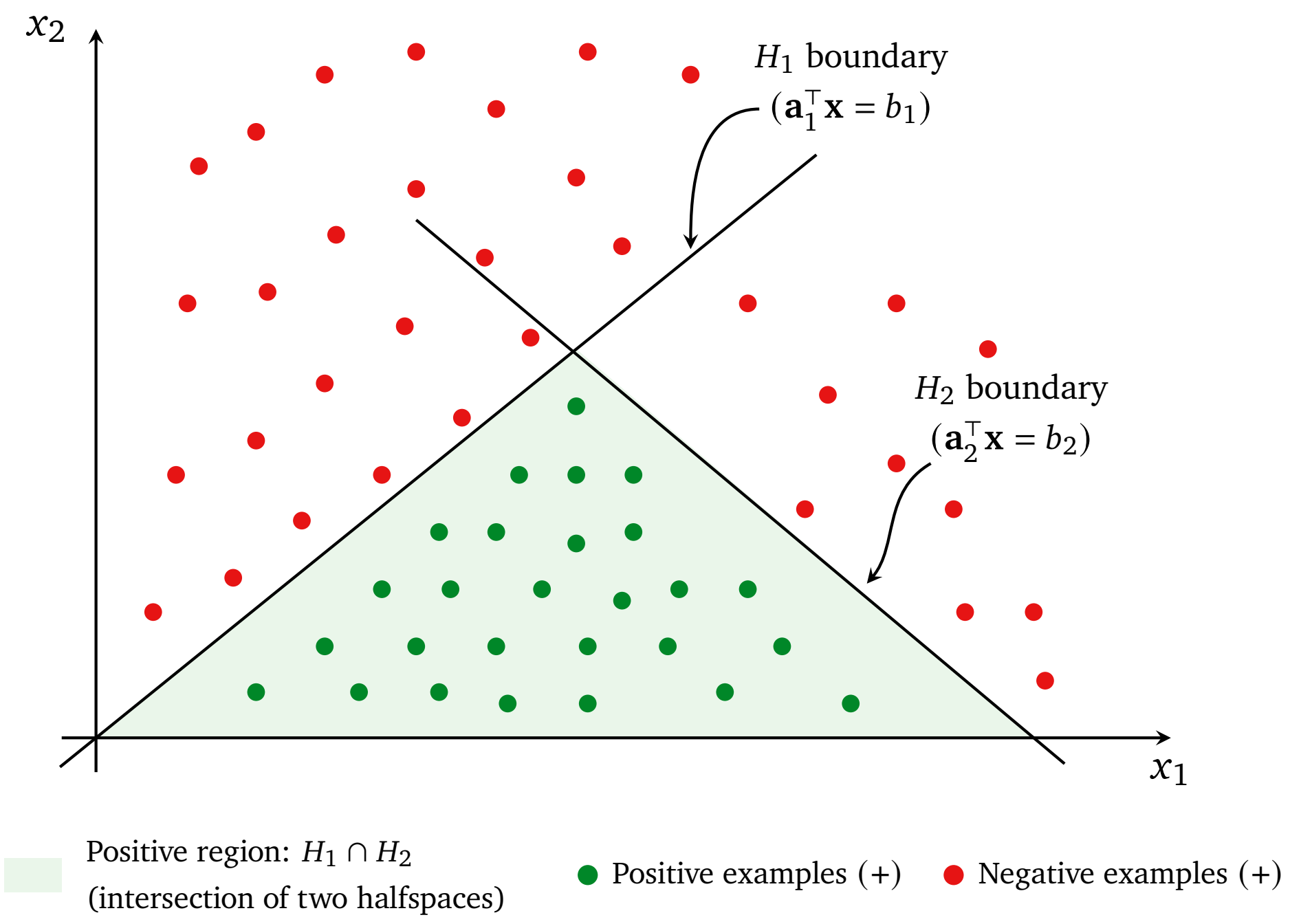

We will denote by $C$ a concept class of $n$-variate Boolean functions, i.e., a class of functions mapping $\{-1,1\}^n$ to $\{-1,1\}$. We usually consider syntactically defined classes of functions such as the class of all $n$-variate
linear threshold functions. For a fixed concept $f \in C$, we are interested in the set
\[
f^{-1}(1) = \{ x \in \{-1,1\}^n : f(x) = 1\}
\]
of examples $x$ labeled positively by the concept $f$. An equivalent way to think of the set $f^{-1}(1)$ comes from an extensive line of works in inductive inference and formal languages (e.g., \citet{gold1967language,angluin1980inductive,kleinberg2024language}): one can think of this set as the elements of a \emph{language} $L$ that is induced by positive assignments of $f$; as an illustration, the reader can think of a particular language $L$ induced by the classifier $f$ as the set of valid Python programs (these are the examples $x$ with $f(x) = 1$, where an example $x$ is a specific valid Python program) and all the points $x$ with $f(x) = -1$ are Python programs that do not compile. {Hence, using the above equivalence, we will associate a class of concepts $C$ with an induced class of languages $\cL = \cL(C)$.} For example, \Cref{fig:placeholder1} shows a language $L$ induced by an intersection of two halfspaces $f$, where $f^{-1}(1)$ corresponds to the green region.

The ``low-complexity'' probability distributions that we are interested in are the uniform distributions on languages induced by concepts $f \in C$. In particular, we want to learn distributions in the class
\[
\{ U_{L}^+ : ~~ L =  f^{-1}(1), f \in C\}\,.
\]
Here, the distribution $U^+_{L}$ on the domain $\{-1,1\}^n$ is the uniform distribution on the ``language'' $L = f^{-1}(1)$, which contains all the positive examples from the concept $f \in C.$ We use the superscript $+$ to emphasize that the distribution is supported on the positive examples.

Now the learning task is defined as follows. 
There is an unknown target distribution $U^+_{L^\star}$ for some unknown true language $L^\star = (f^{\star})^{-1}(1)$ where $f^\star \in C$ and the learning algorithm receives $n$ i.i.d.\ samples from it. The goal is to learn to sample from this distribution (i.e., language) in the following sense:

\begin{definition}
[PAC Learning to Sample (Equivalent model to \cite{de2014learning})] 
\label{def:PAC}
Let $C$ be a concept class of $n$-variate Boolean concepts with induced class of languages $\cL$. 
We say that an algorithm $\mathcal{G}$ $(\varepsilon ,\delta)$-PAC learns to sample\footnote{The task of learning to sample was first proposed in the TCS community by \cite{kearns1994learnability} (as learning a generator) but the exact term was recently proposed by \citet{moitra2021learning}; see also the discussion in \citet{risteski2026computational}. We use the term ``PAC learn to sample'' to clarify that we work in the PAC language generation setting (of \cite{de2014learning}).} $\cL$ if for any parameters $\epsilon,\delta \in (0,1)$ and any target language ${L^\star} \in \cL$, there is a number $n = n_C(\epsilon,\delta)$ such that the algorithm after drawing $n$ i.i.d. samples $S$ from $U^+_{L^\star}$, computes, in time $T_C(\eps,\delta)$, a sampler  $G_S : \{-1,1\}^m \to \{-1,1\}^n$ (with $m = \poly(n))$ so that, with probability $1-\delta$, {it holds that \[\mathrm{TV}(G_S(U), U_{L^\star}^+) \leq \epsilon,\]
where $G_S(U)$ is the output distribution of the sampler conditioned on the training set $S$.
}
\end{definition}

The output of the learning algorithm is a generative model, i.e., a circuit that maps random bits of length $m$ to elements of the Boolean hypercube whose output distribution $G_S(U)$  achieves small total variation distance with respect to the target language. Intuitively, to draw some analogies with the extensive line of works on language generation in the limit, outputting a generator that achieves small total variation distance with the target language $L^\star$ corresponds to finding a language model that (i) is approximately consistent with the target language (it has small hallucination rate \cite{kalai2024calibrated,kleinberg2024language,kalavasis2025limits,kalai2025language,charikar2024exploring}; see e.g., \cref{def:hal}) and (ii) approximately achieves breadth relative to the true language (it is expressive enough to approximately match the richness of $L^\star$ \cite{kalavasis2025limits,kleinberg2024language,kleinberg2026banach,kleinberg2025language,charikar2024exploring,peale2025representative}; see e.g., \Cref{def:breadthBoolean}). For a  further discussion and connections between our definition for PAC learning to sample and other objectives, we refer to \Cref{sec:otherDefs}.

\begin{remark}
[Conditional TV Distance] 
{Importantly in the above definition we condition on the training set and measure the total variation distance after conditioning on the observed data. In other words, the learner cannot simply memorize training examples and generate them, hence enforcing ``creativity'' on the output generative model (this is analogous to the requirement of \citet{kleinberg2024language} in language generation in the limit).} For example, suppose the learner is trained on a set $S$ of size $\poly(n)$ and the target language $L^\star$ is of size $2^{\sqrt{n}}$, and consider a sampler $G_1$ that simply assigns uniform mass on $S$, which has output law $\frac{1}{|S|} \sum_{i \in S} \delta_i$. 
If we measure the TV distance without conditioning on $S$, then the law of $G_1(U)$ would be the same as $U_{L^\star}^+$ over the randomness of $S$, and thus the TV distance is $0$. 
However, the conditional total variation would be $1 - o(1)$ as desired because $|S| \ll |L^\star|$. 
On the other side, let $G_2$ be a sampler that assigns uniform mass on the set $S$ and on a ``creative'' set $T \subset L^\star\setminus S$ with $|T| = 2^{\sqrt{n}}/2$. Then the output law of $G_2$ has conditional total variation $\frac12 + o(1)$ with $U_{L^\star}^+$, {indicating that it is a ``better'' solution than $G_1$.} 
\end{remark}

We also stress that our goal is to obtain a \emph{sampler} for the true language and not merely a density evaluator. Since one of the key aspects of \Cref{def:PAC} is computational efficiency, these two tasks (i.e., density evaluation and generation) are not computationally equivalent in general \cite{kearns1994learnability,naor1996evaluation,chewi2025ddpm,risteski2026computational}.
{Hence, our algorithms should output models that generate efficiently from the target language.}
Having set up the model of interest, we now present our main contributions.

\section{Main Results}

We can now discuss the main question we want to understand in this work and our results. In our model of PAC language generation, our goal is to study the following question:
\begin{center}
    \emph{Given a class $C$ on $\{-1,1\}^n$, what is the complexity of PAC learning to sample $C$?}
\end{center}
We want to understand this question both in terms of samples and in terms of time. If we ignore computation and focus on the sample complexity, then one can prove that there is an algorithm that PAC learns to sample (potentially in exponential time) any $n$-variate Boolean class with sample complexity that scales with its VC dimension. The algorithm is basically the minimally consistent ERM given a set of positive examples and follows from known techniques in the literature of learning from positive and unlabeled examples \cite{denis1998pac} (we know that the marginal distribution is the uniform in our problem). For more details, we refer to \Cref{sec:stat}.

We next discuss our contributions for the more challenging case where the algorithm should run in sub-exponential in the dimension $n$ time. Prior work of \cite{de2014learning} on the problem of \Cref{def:PAC} shows (implicitly) the following elegant efficient reduction in PAC learning to sample a class of concepts $C$:
\begin{center}
\textbf{PAC Learning to Sample} $C$
\\$\downarrow$\\ 
\textbf{Online Learner} for $C$ ~+~ \textbf{Approximate Sampling/Counting Oracle} for $C$ + \textbf{SQ Learner} for $C$ 
\end{center}

Hence the work of \citet{de2014learning} identifies three algorithmic primitives in order to efficiently PAC learn to sample a concept class $C$ of $n$-variate Boolean concepts. In particular, for the reduction to be efficient, the following are needed: 
\begin{enumerate}
    \item The class $C$ should admit a computationally efficient \emph{online} learner (in the sense of Littlestone's mistake bound model \cite{littlestone1988learning}). {This means that in the adversarial setting, there should be an efficient algorithm that competes with an adaptive adversary making a small number of mistakes. We refer to \Cref{def:online} for details.}
    
    \item There must exist efficient approximate sampling and counting oracles for the class $C$; these oracles are fundamental objects in theoretical computer science \cite{jerrum1986random,sinclair1989approximate} with extensive literature and beautiful connections to complexity theory, statistical physics, and many other areas. As an illustration, a sampling oracle for $C$ gets as input a concept $f \in C$ and accuracy $\eps$, and outputs a random variable $X$ whose mass at each point approximates multiplicatively that of the uniform distribution on $f^{-1}(1)$. Similarly, a counting oracle approximates multiplicatively the probability $\Pr_{x \sim \unif(\{\pm 1\}^n)}[f(x)=1]$. {See \Cref{def:apxcount,def:apxsample} for details.} 
    \item The class $C$ should admit an  efficient Statistical Query (SQ) learner, i.e., a learning algorithm that operates in the SQ model of learning \cite{kearns1998efficient}. This restrictive model of learning captures most well-known ML algorithms and is widely used across learning theory (see e.g., \cite{diakonikolas2017statistical,feldman2012complete,feldman2017general,diakonikolas2019recent}). We refer to \Cref{def:sq} for details.  
\end{enumerate}
The algorithm that they propose is \emph{oracle-efficient}: If all these ``oracles'' are computationally tractable (i.e., can be implemented in polynomial time), then there is an algorithm that runs end-to-end in polynomial time, and outputs a sampler for the target language. Our work will explore new oracle-efficient algorithms for PAC learning to sample.

\paragraph{Result I.} One of our motivations is that computationally efficient online learning seems like a very strong requirement for PAC learning to sample. We prove that in fact efficient online learning is not, in general, a requirement for efficient PAC learning to sample:
\begin{inftheorem}
[See \Cref{thm:separation}]
\label{infthm1}
There is a concept class $C$ on $\{-1,1\}^n$ for which online learning is computationally hard assuming the existence of one-way functions but there is a polynomial time algorithm that PAC learns to sample $C$.
\end{inftheorem}

This separation suggests that there might be alternative reductions for PAC learning to sample.
Our separation follows from a modification of a natural concept class based on pseudorandom functions \cite{blum1994separating}, which was recently used to show separations between online with private and replicable PAC learning \cite{bun2020computational,kalavasis2024computational}.

{We mention that the class $C$, {introduced by \citet{blum1994separating}}, is efficiently PAC learnable (in the standard classification sense); also, as we observe, the PAC learner for it \emph{only uses positive examples}. Hence, a natural approach in order to PAC learn to sample is to first learn the target concept from positive examples and then generate from it using rejection sampling. However, rejection sampling for this class is not efficient {since the mass (of some hypotheses) can be exponentially small} and so one cannot immediately get a sampler. We discuss more technical details in \Cref{sec:ows}.}

The existence of strong negative results about efficient online learning, as the ones of \citet{blum1994separating,bun2020computational}, has motivated many ways to bypass hardness in online learning e.g., under models based on smoothed analysis (e.g., \cite{haghtalab2024smoothed}) or by assuming access to weaker oracles. To this end, we settle to consider offline ERM oracles, which have been naturally motivated in various areas of machine learning (e.g., online learning \cite{assos2023online,haghtalab2022oracle}, causal inference \cite{foster2023orthogonal}, and reinforcement learning \cite{foster2020beyond,golowich2024exploration} to name a few). This leads us to our second set of results. 

\paragraph{Results II and III.} 
Our second question is whether access to an offline oracle for learning a concept class $C$ could be useful for PAC learning to sample the set of satisfying assignments of the target concept.
In summary, one of our positive results is a different reduction compared to the one of \cite{de2014learning} that can be informally summarized as follows: First, we replace the online learner with an offline ERM oracle for the class $C$. An ERM oracle for $C$ is defined as an oracle that gets as input a set $S$ of labeled examples from some concept $f^\star\in C$ and outputs a concept in $C$ that achieves zero loss on these examples. The oracle is efficient if it can be implemented in time $\poly(n,|S|).$

Our algorithm below shows that indeed an offline ERM oracle is sufficient for PAC learning to sample. However, there is a trade-off: we should be able to perform approximate sampling (and counting) from distributions that are uniform not only on the sets $f^{-1}(1)$ for $f \in C$ but also \emph{intersections} of these sets {(e.g., if $C$ is the class of LTFs, then we need to sample from intersections of halfspaces).} Our reduction is summarized as follows:

\begin{inftheorem}
[See \Cref{thm:PAC-learn-sample-ERM}]  
{
Let $C$ be a concept class of $n$-variate Boolean concepts with induced class of languages $\cL$.  
Assume that the learning algorithm has i.i.d.\  sample access to the target $U^+_{L^\star}$ and oracle access to:
\begin{enumerate}
    \item an offline ERM oracle for $C$ with runtime $T_{\mathrm{ERM}}$,
    \item an approximate sampling/counting oracle for intersections of concepts in $C$ with runtime $T_{\mathrm{samp}}$,

    \item and an SQ {learning algorithm} for $C$ with runtime $T_{\mathrm
    SQ}$.
\end{enumerate}
Then there is an algorithm that runs in $\poly(n, 1/\epsilon, T_{\mathrm{ERM}}, T_{\mathrm{samp}},T_{\mathrm
    SQ})$ time making {$O(n)$} oracle calls with accuracy $\poly(\epsilon,1/n)$ and PAC learns to sample from the true target language with accuracy $\epsilon$.
\label{infthm2}
}
\end{inftheorem}
\begin{center}
\textbf{PAC Learning to Sample} $C$
\\$\downarrow$\\ 
\textbf{ERM} for $C$ ~+~ \textbf{Approximate Sampling/Counting Oracle} for $\{\bigcap_{i=1}^{k} c_i : c_1,...,c_{k} \in C\}$ + \textbf{SQ Learner} for $C$ 
\end{center}
We note that the number of intersections $k$ that we need is at most $n$. {We give a sketch of our algorithm in \Cref{sec:technical}.
} The above reduction indicates an interesting trade-off between the complexity of the learning oracle (online vs. offline) and the complexity of sampling/counting (concepts in $C$ vs. intersections of concepts in $C$).

We show that this trade-off is necessary in general.
Namely, an offline ERM oracle, approximate sampling and counting oracles for $C$ (instead of intersections of concepts in $C$), and an SQ learner for $C$ do not suffice for efficient PAC learning to sample.

\begin{inftheorem}
[See \Cref{thm:sig-hardness}]
\label{infthm:sig}
Assume that there is a unique signature scheme that is secure against random message attacks.
Then there is a concept class $C$ on $\{-1,1\}^n$ with an efficient ERM oracle, efficient exact sampling and counting oracles, and an efficient SQ learner, such that no polynomial-time algorithm PAC learns to sample $C$.
\end{inftheorem}

In particular, in the reduction of \citet{de2014learning}, the online learner cannot be replaced by an ERM oracle while keeping the sampling and counting oracles for $C$.
The hardness results of \citet{de2014learning} are based on the same assumption.

\paragraph{Result IV.} 
Given the above reduction and the class of \Cref{infthm1}, we can observe a tension between the amount of negative examples required to learn the concept class and the number of intersections one has to be able to generate from. In particular, the two extreme cases are:
\begin{enumerate}
    \item The concept class is offline PAC learnable with only positive examples and hence one only needs a uniform generator for the output of the PAC learner. This was (roughly speaking) for example the case in our \Cref{infthm1}.

    \item The concept class is offline PAC learnable with positive and negative examples and hence our \Cref{infthm2} implies that a more complicated sampling oracle is required that can handle sampling from intersections of concepts in $C$.
\end{enumerate}
{
A natural question is whether the  interpolation between these two extreme cases is possible. Would a class, which can be PAC learned using ``few'' negative examples, require ``few'' intersections in the sampling oracle? We formally prove that such a phenomenon is actually true. To do this, we consider a class of PAC learning algorithms that are robust to deletions of negative examples. This class of algorithms, which we formally call PAC learners robust to $p$-negative filtering ($p$ stands for the fraction of negative examples {not} deleted\footnote{{To give some intuition, consider a labeled dataset with $N_p$ positive and $N_n$ negative examples. Here the PAC learner would achieve an error rate $R(N_p + N_n)$. Consider now an adversary that keeps all the positive examples and with probability $p$ keeps each negative example independently; then the adversary reveals the survival datasets of size (roughly) $N_p + p N_n$}. The learner is robust to the filtering if it can get the same rate $R(N_p + N_n)$ using only the survival dataset.}), interpolates between classical PAC learners (which need both positive and negative examples; here $p=1$) and positive PAC learners (that correspond to the strongest type since they can handle deletion of all negative; here $p=0$). Given access to a PAC learner that is robust to $p$-negative filtering, we prove the following informal result:

\begin{inftheorem}
    \label{infthm3}
    Let $C$ be a concept class of $n$-variate Boolean concepts with induced class of languages $\cL$.
    Assume that the learning algorithm has i.i.d. sample access to the target $U_{L^\star}^+$ and oracle access to:
    \begin{enumerate}
        \item an offline PAC learner for $C$ that is robust to $p$-negative filtering \footnote{We need the error rate of the PAC learner to be linear in $1/N_0$ where $N_0$ is the number of samples before filtering, see \cref{sec:tradeoff} for details.} with runtime $T_{\mathrm{PAC}}$,
        \item an approximate sampling/counting oracle for intersections of $O(n/\log(1/p))$ concepts in $C$\footnote{Here we implicitly assume that our PAC learner is proper. If it outputs concepts in $C'$, we need sampling/counting oracles for $C'$.} with runtime $T_{\mathrm{samp}}$, 
        \item an SQ learning algorithm for $C$ with runtime $T_{\mathrm{SQ}}$,
    \end{enumerate}
    Then there is an algorithm that runs in $\poly(n, 1/\varepsilon, T_{\mathrm{PAC}}, T_{\mathrm{samp}},T_{\mathrm{SQ}})$ time making $\poly(n, 1/\varepsilon )$ oracle calls with accuracy $\poly(\varepsilon ,1/n)$ and PAC learns to sample from the true target language with accuracy $\eps$. 
\end{inftheorem}

It is crucial to observe the regimes of interest:
\begin{enumerate}
    \item When $p \approx 1$, we essentially have a classical PAC learner and so we need a number of intersections of order $n$. This corresponds to our \Cref{infthm2}.
    \item When $p \approx 0$, we essentially have a positive PAC learner and hence we only need to sample/count from $C$ without any intersections. This corresponds to the example class of \Cref{infthm1}.
    \item When $p$ lies between the two extremes, we prove that the number of intersections behaves roughly speaking as $n/\log(1/p)$. The interested reader can find the detailed and exact definitions and intersection complexity in \Cref{sec:tradeoff}.
\end{enumerate}
}

\paragraph{Avoiding Hallucinations} {A natural question is whether one could change our TV distance goal and only focus on avoiding hallucinations without achieving large breadth. A natural approach for that would be again to consider intersection of concepts as follows: Let $S$ be the set of positive examples. We can consider the ERM objective that 
finds all the consistent hypotheses and outputs their intersection. Intuitively, this language should not hallucinate but it will have significantly small breadth. Such approaches have been studied in the literature of learning from positive examples (e.g., \cite{lee2025learning}) but also in the line of works in language generation (e.g.,  \cite{li2024generation}). Since our main focus is on getting total variation guarantees, we leave further investigation of PAC language generation with consistent samplers for future work.
}

\subsection{Technical Overview}
\label{sec:technical}

In this section, we will focus on our new reduction  behind \Cref{infthm2}. For the separation of \Cref{infthm1} we refer to \Cref{sec:ows}, for the hardness result of \Cref{infthm:sig} we refer to \Cref{sec:sig-hardness}, and for the algorithm of \Cref{infthm3}, we refer to \Cref{sec:tradeoff}.
In order to better illustrate our approach in order to get \Cref{infthm2}, it is crucial to mention some potential approaches that fail to solve our problem.

\paragraph{Rejection Sampling} Given i.i.d. samples from $U_{L^\star}^+$, a natural approach is to use tools from Positive and Unlabeled (PU) learning (since the marginal distribution is known to be the uniform) and get a model $\wh h$ with the guarantee that $\Pr_{x\sim \unifcube}[\wh h(x) \neq L^\star(x)] \leq \epsilon.$ Given that, we can simply perform rejection sampling, i.e., generate uniformly random samples and accept $X$ only if $\wh h(X) = 1$. However, this cannot lead to efficiency in general since the mass of the true language can be exponentially small.

\paragraph{Sampling Oracle} A natural idea is to combine the above PU learning approach with a sampling oracle, as the one used by \citet{de2014learning}. The sampling oracle will get as input $\wh h$ and accuracy $\epsilon$ and it will output a sample $X$ with $\mathrm{TV}(\mathrm{Law}(X), \mathrm{Unif}(\wh h^{-1}(1))) \leq \epsilon$. This approach seems valid but the key question is how far is $\mathrm{Unif}(\wh h^{-1}(1))$ from the true target distribution $U_{L^\star}^+?$ 
The answer is that these two distributions can be quite far in total variation and the reason is that PU learning offers \emph{additive} error guarantees.
To see the difference, a PU learner achieves a guarantee of the form:
\[
\sum_{x \in \{-1,1\}^n} \ind\{\wh h(x) \neq L^\star(x)\} \leq \epsilon \cdot 2^n\,.
\]
In contrast, if $G$ is the output distribution induced by $\wh h,$ our guarantee requirement is
\[
\mathrm{TV}(G, U_{L^\star}^+) = 
\sum_{x} |G(x) - U_{L^\star}^+(x)| \leq \epsilon\,.
\]
Hence, the distribution $G$
is evaluated at the \emph{scale of the mass of satisfying assignments of $L^\star$}. To see this, assume that the mass of $L^\star$ is $c \cdot 2^{-n}$ for an absolute constant $c$. Observe that the hypothesis $\wh h = -1$ is a trivial PU learner; however it fails to achieve non-trivial TV distance. In other words, total variation distance learning asks for a \emph{multiplicative} approximation instead of additive \cite{de2014learning}.

\paragraph{Our Approach}
In this section, we will focus on sketching the algorithmic approach behind \Cref{infthm2}. In particular, we will discuss how one can use an offline ERM oracle for the class $C$ together with a sampling and a counting oracle for intersections of concepts in $C$ in order to obtain a hypothesis $h$ that satisfies the following two guarantees:
\begin{enumerate}
    \item The hypothesis achieves almost perfect breadth, in the sense that
    \[
    \Pr_{x \sim U_{L^\star}^+}[h(x) = 1] \geq 1-\epsilon. 
    \]
    \item The hypothesis achieves a non-trivial hallucination rate
    \[
    \Pr_{x \sim \unif(\{\pm1\}^n)}[h(x) =1] \lesssim {\poly(n, \log|C|,1/\delta)} \cdot \Pr_{x \sim \unif(\{\pm1\}^n)}[L^\star(x) =1],
    \]
\end{enumerate}
Given the above predictor, we can make use of an SQ algorithm (in a similar spirit as \cite{de2014learning}) to PAC learn to sample from $C$ (by essentially ``boosting''  its hallucination rate). This is a non-trivial process but we will focus on using our oracles to get the above Items 1 and 2.

First, let us see how to get almost perfect breadth.
Our claim is that \emph{any} hypothesis in $C$ that is \emph{consistent} with a sufficient large set of positive examples drawn from the true target distribution $U_{L^\star}^+$ achieves non-trivial expressivity, i.e., $\Pr_{x \sim U_{L^\star}^+}[f(x) = 1] \ge 1 - \varepsilon$.

\begin{claim}
[Breadth of Consistent Hypotheses]
\label[claim]{claim:breadth}
For any $\varepsilon ,\delta > 0$, if $N_+ \ge \frac{1}{\varepsilon }\log \frac{|C|}{\delta}$, 
then with probability at least $1 -\delta$ over the draw of $x_1^+, \dots, x_{N^+}^+ \overset{\rm i.i.d.}{\sim} U_{L^\star}^+$, 
any function $f\in C$ such that $f(x_i^+)=1$ for all $i \in [N^+]$ 
satisfies $\Pr_{x \sim U_{L^\star}^+}[f(x) = 1] \ge 1 - \varepsilon $. 
\end{claim}

The proof of this claim follows by a union bound, which controls the probability that there exists $f \in C$ that agrees with all the positive examples but assigns probability at most $1-\epsilon$ to the set where $L^\star(x)=1$.

Using the above claim, we can obtain a language $h^0$ that has almost perfect breadth. However, there is no control over the hallucination rate; in particular, $h^0$ can be an extremely large superset of $L^\star.$ The intuition is that in this scenario, we can use the sampling oracle with input $h^0$ to generate a few random examples, which with high probability, they will be \emph{negative} examples for $L^\star$ (they will belong to $h^0 \setminus L^\star).$

The idea then is to use the offline learner that uses the true positive examples from the input and the artificial negative examples in order to get an updated hypothesis $h^1$. The claim is that when we take the intersection $h^1 \cap h^0$, there is a \emph{contraction} of the hallucinations. 

\begin{claim}
[Hallucination Contraction]
\label[claim]{claim:hallucination}
{Fix $i \in [n]$.}
For any $c ,\delta \in (0, 1)$, if $N^- \ge \frac{1}{c} \log \frac{2|C|}{\delta}$, then with probability at least $1 - \delta$ over the draw of $x_1^-,\dots, x_{N^-}^- \overset{\iid}{\sim} \cA^{\bigcap C}_{\rm gen} (i, h^0,\dots, h^{i-1},\delta/2N_-) $, any function $f \in C $ such that $f(x_i^-) = -1$ for all $i \in [N^-]$ satisfies \[\Pr_{x \sim \unifcube}\left[\left(\bigcap_{j=0}^{i-1} h^j \cap f\right)(x) = 1\right] \le c \cdot  \Pr_{x \sim \unifcube}\left[\left(\bigcap_{j=0}^{i-1} h^j\right)(x) = 1\right].\] 
\end{claim}
This means that adding the new $f$ to the previous hypotheses, reduces the ``volume'' of the predicted language by a factor of $c < 1$.  Given the above, our algorithm works iteratively as follows:

\begin{algorithm}[H]
\caption{Informal Version of algorithm for \Cref{infthm2}}
\label{algo:infAlgo}
\SetKwInput{Input}{Input}
\SetKwInput{Oracle}{Oracle Access}
\SetKwInput{Output}{Output}
\Input{Sampler access to $U_{L^\star}^+$, parameters $\varepsilon ,\delta > 0$, the mass of the true language $\alpha$}
\Oracle{ERM oracle, sampling and counting oracle for intersections of $C$}
Generate positive example set $S^+$ of size $N_+ = O(\frac{n}{\varepsilon }\log \frac{|C|}{\delta})$ i.i.d.\ from $U_{L^\star}^+$\;

Run ERM on $S^+$ to get the initial guess $h^0$\;

\For{$i \gets 1$ \KwTo $n$}{
    Generate artificial negative example set $S^-$ of size $N_- = O(\log\frac{n|C|}{\delta})$ i.i.d.\ from the sampling oracle on $\bigcap_{j=0}^{i-1} h^j$\;

    Run ERM on $S^+ \cup S^-$ to get a new guess $h^i$\;
    
    Estimate the mass $\wh \alpha_i$ of guessed language $\bigcap_{j=0}^{i} h^j$ using the counting oracle\;
     
    \If(\tcp*[f]{$\gamma = \delta/(64 n N^-)$}){$\wh  \alpha_i \le \alpha/(\gamma (1+ \varepsilon )^2)$}{

    The mass is small compared to the true one, output the current guess $h=\bigcap_{j=0}^i h^j$\;
        
    }
}
\end{algorithm}

\begin{figure}[!ht]
    \centering
\definecolor{Lgreen}{RGB}{205, 235, 205}
\definecolor{Lgreenborder}{RGB}{35, 120, 35}
\definecolor{Hyellow}{RGB}{254, 243, 205}
\definecolor{Hyellowborder}{RGB}{210, 150, 20}
\definecolor{Hblue}{RGB}{215, 230, 250}
\definecolor{Hblueborder}{RGB}{60, 110, 200}
\definecolor{Ppurple}{RGB}{125, 40, 160}
\definecolor{Rred}{RGB}{220, 35, 35}
\scalebox{0.8}{
\begin{tikzpicture}[x=1cm, y=1cm]

\def\LstarPoly{(4.8, 2.3) -- (6.8, 2.2) -- (5.5, 3.2) -- cycle}
\def\hZeroPoly{(2.1, 1.7) -- (8.8, 1.6) -- (7.8, 5.6) -- (2.5, 5.2) -- cycle}
\def\hOnePoly{(3.8, 1.8) -- (7.3, 1.7) -- (6.3, 4.3) -- (6.0, 4.3) -- cycle}

\newcommand{\greendots}{
  \foreach \p in {(5.3, 2.4), (5.6, 2.9), (6.1, 2.4)} {
    \fill[Lgreenborder] \p circle (1.4pt);
  }
}

\newcommand{\purpledots}[1]{
  \foreach \p in {
(3.3, 1.8),
(4.2, 4.8),
(7.3, 5.1),
(5.0, 3.4),
(6.9, 3.3),
(7.8, 2.8),
(4.1, 2.8),
(8.2, 2.5),
(8.0, 3.3),
(4.4, 3.3),
(3.3, 4.7),
(5.8, 4.5),
(2.3, 1.8),
(7.1, 2.8),
(4.8, 4.1),
(2.6, 3.1),
(2.7, 4.2),
(6.9, 4.0)
  } {
    \fill[#1] \p circle (1.4pt);
  }
}

\newcommand{\reddotsInHzero}{
  \foreach \p in {
(5.0, 5.2),
(3.6, 4.0),
(2.8, 5.0),
(3.8, 5.1),
(4.5, 5.2),
(4.5, 2.3),
(6.6, 2.9),
(4.1, 4.2),
(6.7, 4.5),
(8.0, 3.8),
(8.5, 2.0),
(3.0, 2.5),
(5.6, 5.1),
(7.5, 4.5),
(7.4, 4.0),
(3.5, 2.4),
(8.0, 1.7),
(4.9, 2.9),
(6.3, 3.6),
(4.0, 2.2),
(7.4, 3.5),
(4.7, 4.7),
(5.3, 4.1),
(6.8, 5.3),
(3.0, 3.8),
(7.2, 2.2),
(5.6, 3.6),
(6.0, 4.0),
(6.1, 5.2),
(7.8, 2.2),
(3.3, 3.2),
(5.3, 4.6),
(2.5, 3.6),
(6.6, 2.0),
(2.6, 2.2),
(6.0, 3.2),
(7.8, 4.9),
(4.5, 1.9),
(5.2, 1.5)
  } {
    \fill[Rred] \p circle (1.4pt);
  }
}

\newcommand{\reddotsOutsideHzero}{
  \foreach \p in {
    (8.9, 0.9), (4.6, 5.7), (2.1, 5.3), (0.8, 2.8), (2.5, 5.6), (7.8, 0.9),
    (0.8, 3.9), (9.1, 2.9), (5.5, 0.9), (9.1, 3.8), (1.9, 2.2), (5.6, 5.6),
    (6.6, 0.7), (8.4, 4.1), (1.0, 5.5), (2.0, 3.0), (4.0, 5.6), (1.3, 4.7),
    (1.1, 1.9), (2.0, 3.7), (5.8, 0.5), (0.7, 5.1), (5.9, 1.3), (8.2, 5.4),
    (8.6, 3.6), (4.1, 1.1), (8.7, 4.5), (8.4, 0.9), (1.5, 3.6), (4.5, 0.7),
    (3.4, 5.7), (0.7, 4.5), (5.0, 1.0), (1.6, 4.2), (6.6, 1.2), (0.9, 3.3),
    (1.5, 2.8), (8.4, 4.9), (9.0, 5.7), (0.4, 3.1), (0.6, 1.8), (0.4, 5.8),
    (9.0, 1.7), (8.7, 2.6), (2.1, 4.3), (6.3, 5.7), (9.0, 4.9), (7.0, 0.4),
    (1.8, 4.9), (1.9, 5.8), (1.6, 1.8), (1.5, 5.5), (0.8, 2.3), (7.3, 1.1),
    (7.4, 5.8), (9.0, 2.2), (8.6, 3.1), (5.0, 0.5), (8.3, 0.4), (5.1, 5.7),
    (7.7, 0.4), (8.7, 5.3), (4.0, 0.5)
  } {
    \fill[Rred] \p circle (1.4pt);
  }
}

\begin{scope}[shift={(0,7.0)}]
  \draw[rounded corners=5pt, line width=0.6pt, gray!60] (0,0) rectangle (9.5,6.5);
  \node[font=\small] at (4.75, 6.15) {1. True language $L^\star$ (small region)};

  \path[fill=Lgreen, draw=Lgreenborder, line width=0.8pt] \LstarPoly;

  \node[font=\color{Lgreenborder}] at (6, 2) {$L^\star$};

  \purpledots{Rred}
  \reddotsInHzero
  \reddotsOutsideHzero
  \greendots

  \node[draw=gray!50, fill=white, rounded corners=2pt, inner sep=2pt, anchor=south west] at (0.1, 0.1) {
    \begin{tabular}{@{}l@{\hspace{5pt}}l@{}}
      \tikz\fill[Lgreenborder] (0,0) circle (1.5pt); & \footnotesize Positive (in $L^\star$)\\
      \tikz\fill[Rred] (0,0) circle (1.5pt); & \footnotesize Negative (not in $L^\star$)
    \end{tabular}
  };
\end{scope}

\begin{scope}[shift={(10.2,7.0)}]
  \draw[rounded corners=5pt, line width=0.6pt, gray!60] (0,0) rectangle (9.5,6.5);
  \node[font=\small] at (4.75, 6.15) {2. Larger hypothesis $h^0$ (superset of $L^\star$)};

  \fill[Hyellow] \hZeroPoly;
  \draw[line width=0.8pt, Hyellowborder] \hZeroPoly;
  \path[fill=Lgreen, draw=Lgreenborder, line width=0.8pt] \LstarPoly;

  \node[font=\color{Lgreenborder}] at (6, 2) {$L^\star$};
  \node[font=\color{Hyellowborder}] at (8.1, 5.8) {$h^0$};

  \purpledots{Rred}
  \reddotsInHzero
  \reddotsOutsideHzero
  \greendots

  \node[draw=gray!50, fill=white, rounded corners=2pt, inner sep=2pt, anchor=south west] at (0.1, 0.1) {
    \begin{tabular}{@{}l@{\hspace{5pt}}l@{}}
      \tikz\fill[Lgreenborder] (0,0) circle (1.5pt); & \footnotesize Positive (in $L^\star$)\\
      \tikz\fill[Rred] (0,0) circle (1.5pt); & \footnotesize Negative (not in $L^\star$)
    \end{tabular}
  };
\end{scope}

\begin{scope}[shift={(0,0)}]
  \draw[rounded corners=5pt, line width=0.6pt, gray!60] (0,0) rectangle (9.5,6.5);
  \node[font=\small] at (4.75, 6.15) {3. Mark artificial negative examples in $h^0 \setminus L^\star$};

  \fill[Hyellow] \hZeroPoly;
  \draw[line width=0.8pt, Hyellowborder] \hZeroPoly;
  \path[fill=Lgreen, draw=Lgreenborder, line width=0.8pt] \LstarPoly;

  \node[font=\color{Lgreenborder}] at (6, 2) {$L^\star$};
  \node[font=\color{Hyellowborder}] at (8.1, 5.8) {$h^0$};

  \purpledots{Ppurple}
  \reddotsInHzero
  \reddotsOutsideHzero
  \greendots

 \node[draw=gray!50, fill=white, rounded corners=2pt, inner sep=2pt, anchor=south west] at (0.1, 0.1) {
    \begin{tabular}{@{}l@{\hspace{5pt}}l@{}}
      \tikz\fill[Lgreenborder] (0,0) circle (1.5pt); & \footnotesize Positive (in $L^\star$)\\[-0.5ex]
      \tikz\fill[Rred] (0,0) circle (1.5pt); & \footnotesize Negative (not in $L^\star$) \\[-0.5ex]
      \tikz\fill[Ppurple] (0,0) circle (1.5pt); & \footnotesize Artificial negative (in $h^0 \setminus L^\star$)
    \end{tabular}
  };
\end{scope}

\begin{scope}[shift={(10.2,0)}]
  \draw[rounded corners=5pt, line width=0.6pt, gray!60] (0,0) rectangle (9.5,6.5);
  \node[font=\small] at (4.75, 6.15) {4. New hypothesis $h^1$ (contains $L^\star$, excludes purple points)};

  \fill[Hyellow] \hZeroPoly;
  \draw[line width=0.8pt, Hyellowborder] \hZeroPoly;

  \path[fill=Hblue, draw=Hblueborder, line width=0.8pt] \hOnePoly;

  \path[fill=Lgreen, draw=Lgreenborder, line width=0.8pt] \LstarPoly;

  \node[font=\color{Lgreenborder}] at (6, 2) {$L^\star$};
  \node[font=\color{Hyellowborder}] at (8.1, 5.8) {$h^0$};
  \node[font=\color{Hblueborder}] at (6.4, 4.6) {$h^1$};

  \purpledots{Ppurple}
  \reddotsInHzero
  \reddotsOutsideHzero
  \greendots

  \node[draw=gray!50, fill=white, rounded corners=2pt, inner sep=2pt, anchor=south west] at (0.1, 0.1) {
    \begin{tabular}{@{}l@{\hspace{5pt}}l@{}}
      \tikz\fill[Lgreenborder] (0,0) circle (1.5pt); & \footnotesize Positive (in $L^\star$)\\[-0.5ex]
      \tikz\fill[Rred] (0,0) circle (1.5pt); & \footnotesize Negative (not in $L^\star$) \\[-0.5ex]
      \tikz\fill[Ppurple] (0,0) circle (1.5pt); & \footnotesize Artificial negative (in $h^0 \setminus L^\star$)
    \end{tabular}
  };
\end{scope}
\end{tikzpicture}}
    \captionsetup{margin=0.5cm}
    \caption{Cartoon sketch of our algorithm. Imagine all the dots form the Boolean cube $\{-1, 1\}^n$. In Fig.\ 1, we depict with green the target language $L^\star$ (which is a polytope). In Fig.\ 2, our algorithm, uses positive examples to find a initial guess $h^0$, which has good breadth but hallucinates a lot (yellow region). In Fig.\ 3, we use the sampling oracle to get positive examples from $h^0$ (purple dots). Because of the volume of the yellow set is large compared to the green one, they will be with high probability negative examples for $L^\star$ (outside of the green region). In Fig.\ 4 we run ERM on the true positive examples and the artificial negative examples (purple dots) to get a new hypothesis $h^1$ that still covers most of the true language but achieves a lower amount of hallucinations (blue region).}
    \label{fig:placeholder}
\end{figure}
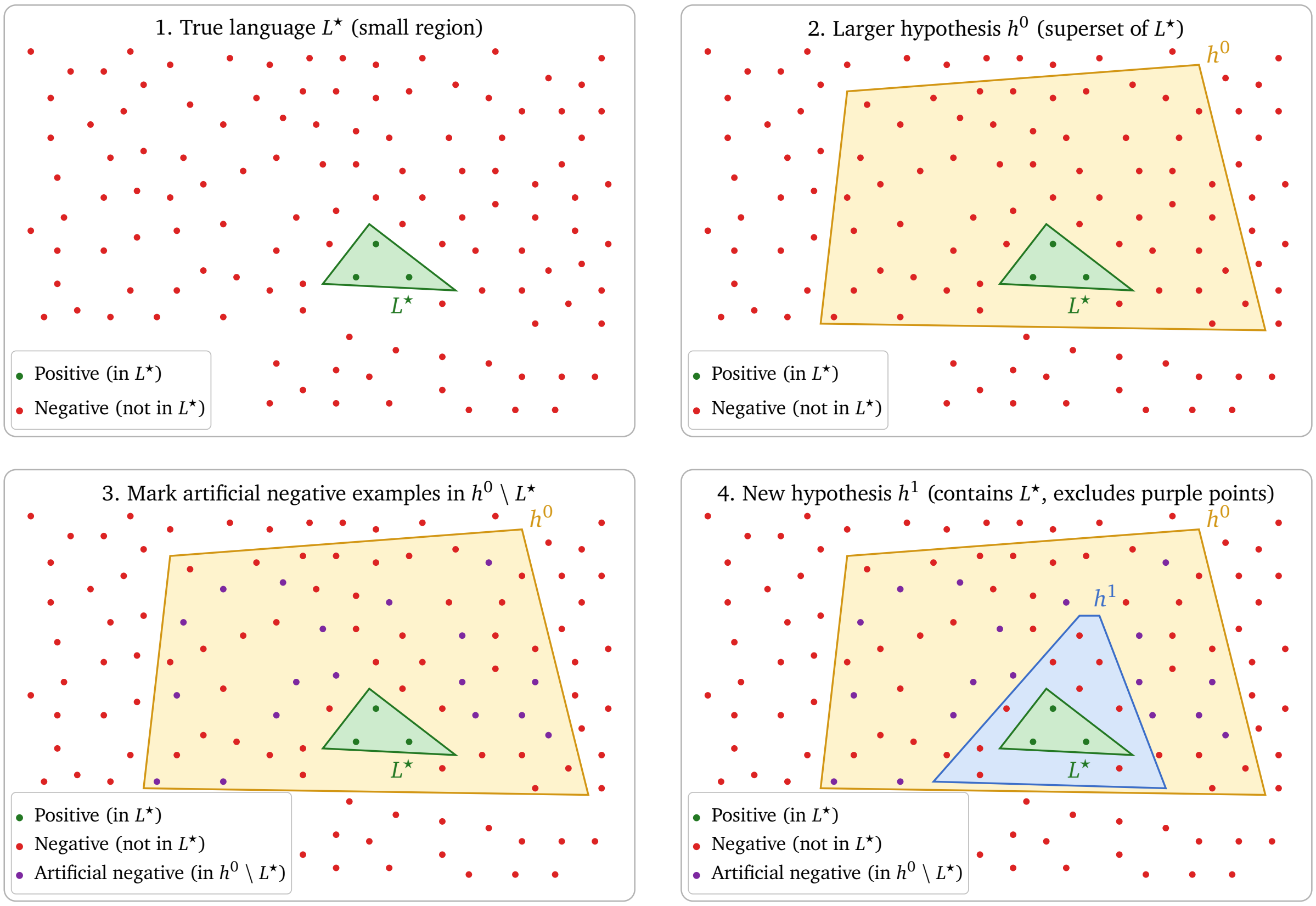

First, we have the following claim.
\begin{claim}
\label[claim]{claim:negative}
    For any iteration $i,$ all the $N_-$ samples are negative for $L^\star$ with high probability.
\end{claim}
This follows because the mass $\alpha_{i-1}$ of the guess language $g_{i-1} = \bigcap_{j=0}^{i-1} h^j$ is at least $\alpha/\gamma$, where $\alpha$ is the given mass of the true language\footnote{For the sketch, we will assume that we have access to the mass $\alpha$. Our algorithm does not need to know the true mass and can get an estimation of it using doubling. We discuss that in detail in the formal proof.}. 

Because of the \cref{claim:negative}, for any $i \in [n]$, with high probability, the output $h^i$ of the ERM oracle
\[h^i = \argmin_{f\in C} \sum_{j=1}^{N_+} \ind[f(x_j^+) \ne 1] + \sum_{j=1}^{N_-} \ind[f(x_j^-) \ne -1]\]
satisfies $h^i(x_j^+) = 1$, $j\in [N^+]$ and $h^i(x_j^-) = -1$, $j\in [N^-]$, since $L^\star$ achieves this minimum.

By \cref{claim:breadth}, any $f \in C$ with $f(x_i^+) = 1$ for all $i \in [N^+]$ has 
$\Pr_{x \sim U_{L^\star}^+}[f(x)=1] \gtrsim 1 - \epsilon/n$
(i.e., sufficiently large breadth).

By \cref{claim:hallucination}, any $f \in C$ with $f(x_i^-) = -1$ for all $i \in [N^-]$, we get the following volume contraction:
\[
\Pr_{x \sim \unif}\left[
\left(\bigcap\nolimits_{j=0}^{i-1} h^j \cap f\right)(x)=1\right]
\leq 
1/2 
\cdot 
\Pr_{x \sim \unif}\left[
\left(\bigcap\nolimits_{j=0}^{i-1} h^j\right)(x)=1\right]
\]
where we picked the contraction parameter $c =1/2$. 
If we condition that all the above events hold, then we can control the volume of the $i$-th predicted language as
\[
\Pr_{x \sim \unif}\left[\left(\bigcap\nolimits_{j=0}^i h^j\right)(x) = 1\right] \leq 2^{-i}\,.
\]
This means that $\wh \alpha_i \leq (1+\epsilon) 2^{-i}$, using the approximation guarantees of the counting oracle. Since when $i = n,$ we get that $\wh \alpha_n \lesssim \alpha/\gamma$, there should exist an iteration where our algorithm will stop and output $h = \bigcap_{j = 0}^{\wh i} h^j$ for some index $\wh i.$ For this hypothesis $h,$ we get that
\[
\Pr_{x \sim \unif}[h(x) = 1] \leq \alpha/\gamma = \Pr_{x \sim \unif}[L^\star(x) =1 ]/ \gamma
\]
and 
\[
\Pr_{x \sim U_{L^\star}^+}[h(x) \neq 1] \leq \epsilon
\]
by union bound. This implies the two desired properties for our hypothesis, since $\gamma = \poly(n, \log|C|, 1/\delta).$

Up to this point, we have found a hypothesis $h$ that achieves almost perfect breadth and has a hallucination rate
\[
\Pr_{x \sim \unif}[h(x) = 1] \lesssim \poly(n,\log|C|,1/\delta) \cdot \Pr_{x \sim \unif}[L^\star(x) =1 ]\,.
\]
This roughly speaking means that our language $h$ essentially covers $L^\star$ but its volume is at most polynomially larger than the volume of the true language, leading to hallucinations. However, we have already made some progress since we started with a language $h^0$ that could hallucinate in a much larger rate. In order to further reduce hallucinations, we follow a technique introduced in \citet{de2014learning} which relies on \citet{denis2005learning}. 
The idea is that, given the hypothesis $h$ above which has $1/\gamma$ hallucination rate, samples from the true language and parameters $\epsilon, \wh \alpha$, we can run the SQ algorithm for the class $C$ under the distribution $U_h^+$, as long as $\wh \alpha$ is \emph{multiplicatively} close to $\Pr_{x \sim \unif}[L^\star(x) =1 ]$, to get a distribution $\wh D$ that is $\epsilon$-close in total variation to the true language (see the algorithm in \cite[Appendix C.2]{de2014learning}).
The reason this is possible is that for a statistical query $q$ under distribution $D$ we can write
\[
\E_{x \sim D}[q(x, L^\star(x))]
= \E_{x \sim D}[q(x,-1)] + \Pr_{x \sim D}[L^\star(x)=1] \cdot \E_{x \sim D_{L^\star}^+}[q(x,1) - q(x,-1)],
\]
where $D_{L^\star}^+$ is the distribution $D$ conditioning on $L^\star(x)=1$. 
The SQ learning algorithm will run under distribution $D=U_h^+$. 
One can estimate $\E_{x \sim D}[q(x,-1)]$ using the uniform generation oracle for $h$. Also given $\wh \alpha$ and because $h$ achieves almost perfect breadth, $\Pr_{x \sim D}[L^\star(x) = 1] \approx \wh \alpha / \Pr_{x\sim \unif}[h(x)=1]$, and $D_{L^\star}^+$ is close to $U_{L^\star}^+$.
Thus, one can simulate statistical queries given samples from $U_{L^\star}^+$. 
Set the accuracy parameter of the SQ learning algorithm to be $\varepsilon  \gamma$ (as $\gamma$ is inverse polynomial), then the output of it $\wh h$ satisfies that $\Pr_{x \sim U_{h}^+}[\wh h(x) \neq L^\star(x)] \leq \epsilon\gamma$. 
Because $h$ achieves almost perfect breadth, this translates to a \emph{multiplicative} error 
\[\Pr_{x\sim \unif}[\wh h(x) \ne L^\star(x)] \approx \Pr_{x \sim U_h^+}[\wh h(x) \ne L^\star(x)] \Pr_{x\sim\unif}[h(x)=1] \le \varepsilon \gamma \lesssim \wh\alpha \varepsilon. \]
The final sampler does rejection sampling according to $\wh h$ on draws from the approximate uniform generator for $h$. The arguments of \citet{de2014learning} prove that this sampler is $\epsilon$-close in total variation to the true language (see Lemma 35 in \cite{de2014learning}). To complete the proof, we need an accurate guess of the true mass. For that, we need to enumerate the guesses $\wh \alpha = (1+ \varepsilon )^i/2^n$ for $i = 0,1,\dots, O(n/\varepsilon )$, so that one of them in the enumeration satisfies $\alpha \le \wh \alpha \le (1 + \varepsilon )\alpha$, where $\alpha$ is the true mass $\Pr_{U}[L^\star(x) =1 ]$. Under each guess, the SQ algorithm gives a different hypothesis and then \cite{de2014learning} runs a hypothesis testing process to select a good concept from the candidates. The details appear in \cite[Appendix C]{de2014learning}. We refer the reader to \Cref{sec:densifier} for the complete proof of how to compute a densifier given the ERM oracle and sampling/counting oracles for intersections of $C$.

\section{Related Work}
The success of modern generative models has motivated an extensive line of works that aim to provide a theoretical abstract understanding on \emph{learning to generate}. Learning to sample is a classical topic in computational learning theory that has been studied from many different works and perspectives \cite{kearns1994learnability,naor1996evaluation,chen2022learning,chen2024learning,de2014learning,anderson2013efficient,canonne2020learning,tewari2025if,xiao2010learning,moitra2021learning,chewi2025ddpm,risteski2026computational,chen2022sampling,lee2023convergence,kim2025train}.
 We review recent and old works in the topic that are closely related to our paper.

\paragraph{Learning from Satisfying Assignments} 
Our work is mostly related with the model of \citet{de2014learning} and \citet{canonne2020learning}. In the seminal work of \citet{de2014learning}, it was proved that  linear threshold functions and DNF formulas admit efficient PAC learning to sample algorithms. For halfspaces, the algorithm runs in $\poly(n,1/\epsilon)$ time while for DNFs, it runs in quasi-polynomial time. Moreover, it was proved that even degree-2 PTFs are cryptographically hard to PAC learn to sample. Our work is motivated by this framework and explores the use of offline oracles in the context of PAC learning to sample. The follow-up work of \citet{canonne2020learning} studied the PAC learning to sample question in the continuous domain and designed  algorithms for LTFs under log-concavity of the marginal distribution. Moreover, they proved that when the marginal is Gaussian, degree-2 PTFs are learnable in this model while degree-4 PTFs are not.
Recently, the work of \citet{feng2026learning} studied learning CNFs from satisfying assignments. They focused on \emph{exact} learning of the target formula from uniformly random solutions under LLL-type conditions (or under some average case instances), improving the sample complexity of Valiant's elimination algorithm \cite{valiant1984theory} for these subclasses.  

\paragraph{The work of \citet{kleinberg2024language}}  The starting point of our work is the model of language generation in the limit. Language generation in the limit has been introduced by \citet{kleinberg2024language}, as a natural (yet surprising) counterpart to Gold's classical model of language identification in the limit \cite{gold1967language}.
A crucial aspect of this model is the \emph{in the limit} property: after a finite amount of examples, the generator stops making mistakes and becomes consistent with the true language. 
By now, there is an extensive line of works in this model; see the {recent \href{https://languagegeneration.github.io/}{COLT 2025 Workshop} for an overview of the area}.
Our motivation for this work is to revisit language generation and adopt a PAC learning perspective by taking \emph{computation} into account, a point that was also raised as an exciting direction in \cite{panigrahi2026hallucination}. Also, we shortly mention that recent works also study other resource-constrained language generation models such as space-efficient generators \cite{kleinberg2026language,flammarion2026space}.

\paragraph{Uniform Generation} After the introduction of language generation in the limit, there are many works that studied the question of \emph{uniform} generation\footnote{This term should not be confused with the uniform distribution that we consider as the marginal of our examples.}, which requires that the generator terminates after a fixed number of training steps and this stopping time is uniform across all target languages \cite{arenas2025language,li2024generation,charikar2024exploring}. By definition, finite concept classes (as the ones we consider in the Boolean domain) admit uniform generators. In our work, we also consider uniformity (through the lens of PAC learnability) by studying both statistical and computational aspects of the problem. It is worth mentioning that the work of \citet{arenas2025language} focuses on the sample complexity of language generation in the limit for natural finite families with a perspective that shares similarities to our work: they also identify that the results of \citet{kleinberg2024language} are qualitative (since the algorithm eventually converges) and they want to understand whether finite sample complexity bounds are possible. Interestingly, they show  sample complexity lower bounds for context-free and regular
languages, and even for strict subclasses
such as locally threshold testable languages.

\paragraph{PU Learning} As already discussed in the seminal work of \cite{de2014learning}, our model is different from the classical learning problem of learning with positive unlabeled examples because we aim to get a \emph{multiplicative} approximation to the optimal classifier, while these works (e.g., \cite{denis1998pac,lee2025learning,mansouri2026learning}) focus on \emph{additive} approximations (which is the standard notion in PAC learning). In other words, additive and multiplicative approximations coincide when the target language has constant volume compared to the Boolean domain.  However, as the volume of the true language decreases, then efficient multiplicative approximation becomes much more challenging since the learner has to adapt to the unknown language's mass.  
{To see this, in classical PU learning under the uniform marginal, the goal is to find a hypothesis $\wh f$ that approximates $f^\star$ in target accuracy $\epsilon$ in the sense that $\Pr_{U}[\wh f(x) \neq f^\star(x)] \leq \eps$. This means that the constant $-1$ hypothesis is good for any $f^\star$ with $|\{x : f^\star(x) = 1\}| \leq 2^n \cdot \epsilon$. This means that when the language has small volume, the problem becomes easy; this is not true in total variation learning since the guarantee is measured ``at the scale'' of $|\{x : f^\star(x) = 1\}|$. Also, natural algorithmic approaches based on polynomial approximation for PU learning \cite{lee2025learning} fall short in our setting because the degree of the polynomial depends on the mass of the true language.}
 Learning from positive examples has recently received extensive study in many settings beyond language identification and generation, mainly in the area of truncated statistics (in continuous \cite{daskalakis2018efficient,,lee2024efficient,kouridakis2026linear,daskalakis2019computationally,nagarajan2020analysis,kontonis2019efficient,zampetakis2026private,diakonikolas2024statistical,,wang2026mean,de2023testing,de2024detecting,daskalakis2021statistical} and discrete settings \cite{fotakis2020efficient,chauhan2025learning,galanis2025one,galanis2024learning,chauhan2026efficient}). It would be interesting to see if tools from these lines of work could be useful for PAC language generation.

\subsection{AI Statement}
All ideas and technical contributions in this paper were human-generated (from November 2025 to April 2026), except for the construction of \cref{thm:sig-hardness}, which was found using GPT-6 Astra in October 2026. 
\Cref{fig:placeholder,fig:placeholder1} were generated by AI models.

\section{Statistical Complexity of PAC Learning to Sample}
\label{sec:stat}
Our approach for the statistical complexity of PAC learning to sample (which does not work when one puts computational constraints on the problem) is that one can employ classical results from classical PAC learning with positive and unlabeled examples \cite{denis1998pac}.
\begin{theorem}
[Sample Complexity of PAC Learning to Sample]
\label{thm:stat}
Let $C$ be a concept class of $n$-variate Boolean concepts with induced class of languages $\cL$. 
There is an algorithm that given as input any accuracy parameter $\epsilon \in (0,1)$ and $n \geq \wt{\Omega}(\mathrm{VC}(C)/\epsilon)$ i.i.d. samples $S$ from $U_{L^\star}^{+}$ with $L^\star \in \cL$, outputs a sampler $G_S$ that PAC learns to sample $L^\star$, i.e., with probability $99\%$ , it holds that
    \[
    \mathrm{TV}(G_S(U), U_{L^\star}^+) \leq \epsilon\,.
    \]
\end{theorem}
This means that PAC learnability of the target concept class is also sufficient for designing a statistical model that fits the density of $U^+_{L^\star}$ for any $L^\star \in \mathcal{L}$ using positive examples. 
We add a proof for completeness in \Cref{sec:proof-stat}.

\section{Algorithms for PAC Learning to Sample}

In this section, we are interested in understanding when we can make the above statistical results algorithmic. First, we present the implicit reduction of \citet{de2014learning} that provides an initial answer to this question. Then, we discuss our algorithms for PAC learning to sample concept classes $C$ on the Boolean hypercube. We start by reviewing the reduction of \cite{de2014learning}.

\subsection{The Result of \cite{de2014learning}}
\label{sec:resultDDS}
Our work is inspired by the work of \citet{de2014learning} which implicitly shows an elegant algorithm for PAC learning to sample a class $C$ using sample access to the true language as well as {access to the following algorithmic primitives}.

{
The first ingredient for the reduction of \citet{de2014learning} is an online learner $\mathcal A_{\onl}^C$ which controls the mistake bound in the realizable online learning model of \citet{littlestone1988learning}. 
\begin{definition}
[Online Learning Algorithm \cite{littlestone1988learning}]
\label{def:online}
Let $C$ be a class of Boolean functions over $\{-1,1\}^n$, and let
$f \in C$ be the unknown target concept. In the online game, learning
proceeds in stages. At stage $t$, the learner is given a point
\[
    x_t \in \{-1,1\}^n
\]
and must predict the label $f(x_t)$. After making its prediction
$\widehat{y}_t$, the learner is given the correct label
\[
    y_t = f(x_t).
\]
An online learning algorithm $\cA^C_{\onl}$ learns $C$ with mistake bound $M$ if, for every
$f \in C$ and every sequence of examples
\[
    x_1,x_2,\ldots 
\]
labeled consistently with $f$, the total number of mistakes satisfies
\[
    \left|\{t : \widehat{y}_t \neq f(x_t)\}\right| \le M.
\]
We say that the online learning algorithm is efficient if the algorithm makes at most $M = \poly(n)$ mistakes and each prediction is computed in $\poly(n)$ time. 
\end{definition}
As an illustration, linear thresholds functions on $\{-1,1\}^n$ admit an efficient online learning algorithm with mistake bound $\Theta(n^2 \log n)$ and $\poly(n)$ time, thanks to a classical result of \citet{maass1994fast}.

Apart from the online learner, the reduction of \citet{de2014learning} requires access to some classical oracles from complexity theory: an approximate counting and an approximate (uniform) sampling oracle for the concept class $C$ \cite{jerrum1986random,sinclair1989approximate}. 

An approximate counting oracle for $C$ gets as input the description of a concept $f \in C$ and outputs a \emph{multiplicative}
approximation for its mass on the satisfying assignments, i.e., the value $\Pr_{x\sim\unifcube}[f(x) = 1]$. Typically, we define it as follows:
} 

\begin{definition}[Approximate Counting]
\label{def:apxcount}
Let $C$ be a concept class of $n$-variate Boolean functions. A randomized algorithm $\cA^C_{\cnt}$ is an approximate counting algorithm for $C$ if given any $f\in C$ and $ \varepsilon,\delta > 0$, it outputs a value $\wh  \alpha$ in time $T_{\cnt}(n, 1/ \varepsilon ,1/\delta)$ such that with probability at least $1 - \delta$, \[\frac{1}{1+ \varepsilon } \Pr_{x\sim \unifcube}[f(x) = 1] \le \wh  \alpha \le (1+ \varepsilon) \Pr_{x\sim \unifcube}[f(x) = 1]. \]
{We say that the approximate counting algorithm is efficient if $T_{\cnt}(n,1/\varepsilon ,1/\delta) = \poly(n, 1/ \varepsilon , \log(1/ \delta))$. }
\end{definition}
Many Boolean concept classes, such as halfspaces, admit approximate counting oracles that are efficient, i.e, they run in time $\poly(n,1/\epsilon)$ \cite{dyer2003approximate}. However, there are concept classes such as PTFs where multiplicative approximation is computationally intractable (because even deciding if $\Pr_{x\sim \unifcube}[f(x) = 1] \neq 0$ is out of reach assuming $\mathsf{NP} \neq \mathsf{RP}$ by a standard reduction from $\mathsf{MaxCut}$ even for degree 2) (see e.g., the discussion in \cite{de2014efficient}).

Similarly, an approximate uniform generation oracle for $C$ gets as input the description of a concept $f \in C$ and outputs a sample whose mass at each point approximates multiplicatively that of the uniform distribution $\unif(f^{-1}(1))$ over the satisfying assignments of $f$.
Note that the approximate counting for LTFs can be used to get approximate generation \cite{dyer2003approximate} thanks to self-reducibility and classical sampling-to-counting reductions \cite{sinclair1989approximate,jerrum1986random}.

\begin{definition}[Approximate Uniform Generation]
\label{def:apxsample}
Let $C$ be a concept class of $n$-variate Boolean functions. A randomized algorithm $\cA^C_{\gen}$ is an approximate uniform generation algorithm for $C$ if for any $f\in C$ and $ \delta > 0$, there is a distribution $D$ supported on $f^{-1}(1)$ such that for each $x \in f^{-1}(1)$,
\[
\frac{1}{1+\delta} \cdot \frac{1}{|f^{-1}(1)|} \le D(x) \le (1+\delta) \cdot \frac{1}{|f^{-1}(1)|},
\]
and, given $f$ and $\delta$, the algorithm runs in time $T_{\gen}(n,1/\delta)$ and either outputs a point $x \in f^{-1}(1)$ distributed according to $D$, or reports failure.
Moreover, the probability of failure is at most $\delta/2$.
{We say that the approximate uniform generation algorithm is efficient if $T_{\gen}(n, 1/\delta) = \poly(n , 1/ \delta)$. }
\end{definition}

Finally, the last ingredient for the reduction of \citet{de2014learning} is {a statistical query  (SQ) learning algorithm \cite{kearns1998efficient}, where the learning algorithm is only allowed to obtain estimates of some statistical properties of the samples but cannot directly access the individual samples. 
Formally, in the SQ model, the learning algorithm has access to a statistical query (SQ) oracle. }
The SQ oracle for a distribution on $\{-1,1\}^n$ and a concept $f \in C$ gets as input a query function $q$ (that maps labeled examples $(x, f(x))$ to $[-1,1]$) and a tolerance $\tau$, and returns a value
$v$ that $\tau$-approximates the $\E_{x \sim D}[q(x, f(x))]$.

\begin{definition}
[Statistical Query Oracle \cite{kearns1998efficient}]
Let $D$ be a distribution over $\{-1,1\}^n$, and let $f \in C$ be the
target concept. A statistical query oracle
$
    \mathrm{STAT}(f,D)
$
takes as input a pair $(q,\tau)$, where
\[
    q : \{-1,1\}^n \times \{-1,1\} \to [-1,1]
\]
is a query function and $\tau > 0$ is a tolerance parameter. The oracle
returns a value $v$ satisfying
\[
    \left|
        \mathbb{E}_{x \sim D}\big[q(x,f(x))\big] - v
    \right|
    \le \tau\,.
\]
\end{definition}
Given the SQ oracle, an SQ learning algorithm is defined as follows:
{
\begin{definition}[Statistical Query Learning Algorithm]
\label{def:sq}
Let $C$ be a concept class of $n$-variate Boolean functions. A randomized algorithm $\cA_{\rm SQ}^{C}$ is an SQ learning algorithm if for any distribution $D$ over $\{\pm  1\}^n$ and any $f \in C$, given $\varepsilon ,\delta > 0$ and query access to the oracle ${\rm STAT}(f, D)$, it outputs a hypothesis $h: \{\pm 1\}^n \to \{\pm 1\}$ with probability at least $1-\delta$ such that $\Pr_{x\sim D}[f(x) \ne h(x)] \le \varepsilon $. 
Let $T_{\rm SQ}(n, \varepsilon ,\delta)$ be the running time of $\cA_{\rm SQ}^C$ (assuming each oracle query takes unit time), and $\tau(n, \varepsilon ,\delta)$ be the minimum tolerance parameter over all the oracle queries from $\cA_{\rm SQ}^C$. 
Then $\cA_{\rm SQ}^C$ is called an \emph{efficient} SQ learning algorithm if $T_{\rm SQ}(n, \varepsilon ,\delta)$ and $\tau(n, \varepsilon ,\delta)^{-1}$ are both at most $\poly(n, 1/\varepsilon ,1/\delta)$. 
\end{definition}
}

It is well-known that most learning algorithms belong to the class SQ, with notable exceptions the Gaussian elimination algorithm for learning parities in the realizable setting \cite{reyzin2020statistical,blum2003noise,blum1998polynomial,feldman2017general}.

Given these {algorithmic primitives}, \citet{de2014learning} implicitly prove the following learning result.

\begin{theorem}
    [Implicit in \cite{de2014learning}]
    \label{thm:dds}
    Let $C$ be a concept class of $n$-variate Boolean concepts with induced class of languages $\cL$. {Then for any $\eps, \delta$}, there is an algorithm that PAC learns to sample $C$ {up to $\varepsilon $ error with $1- \delta$ probability}
    assuming i.i.d. sample access to the true language access and access to an online learning algorithm for $C$,
    an approximate uniform generation and counting oracle of $C$, and an SQ learning algorithm for $C$. 
    Moreover, the algorithm 
    \begin{enumerate}
        \item uses $m = \poly(n, 1/\eps, \log(1/\delta))$ i.i.d. samples from the true target distribution,
        \item plays at most $\poly(n, 1/\epsilon)$ online games ({with} the online learning algorithm); in each game, {the online learner} makes at most
        $\poly(n)$ mistakes and runs in $T_{\onl}(n)$ time,
        \item makes $\poly(n, 1/\epsilon)$ queries to {the} approximate uniform generation oracle for $C$ with accuracy {$\delta'$} and runtime {$T_{\gen}(n,1/\delta' )$}, 
        \item makes $\poly(n,1/\epsilon)$ queries to the approximate counting oracle for $C$ with accuracy $\epsilon'$, confidence $\delta'$, and runtime $T_{\cnt}(n, 1/\epsilon', 1/\delta')$,
        \item 
        makes $\poly(n, 1/\varepsilon )$ calls to the SQ learning algorithm for $C$ with accuracy $\epsilon'$ and confidence $\delta'$, that runs in $t_1(n, 1/\varepsilon' , 1/\delta')$ time and each SQ query takes $t_2(n)$ time to evaluate with tolerance $\tau(n, 1/\epsilon')$,
    \end{enumerate}
    {where $\varepsilon ',\delta' = \poly(n,1/\varepsilon ,1/\delta)$}. 
    The algorithm runs in time 
    \[
\poly(n, 1/\eps, \log(1/\delta),
T_{\onl}(n),
T_{\gen}(n, 1/\delta'),
T_{\cnt}(n,1/\epsilon',1/\delta'), 
t_1(n,1/\epsilon',1/\delta'), t_2(n) ,\tau^{-1}(n, 1/\varepsilon' )
    ).
    \]
\end{theorem}

As an application of this reduction, they prove that LTFs can be PAC learned to sample in polynomial time since as we discussed all these oracles can be implemented efficiently for halfspaces on the Boolean domain. Moreover, they give a quasi-polynomial time algorithm for learning DNFs. 

\subsection{Computational Separation}
\label{sec:ows}

The reduction of \cite{de2014learning} for PAC learning to sample crucially relies on the existence of an \emph{efficient online} learning algorithm for the class $C$. A natural question is to ask whether such an oracle is required for PAC learning to sample in polynomial time. In this section, we show that the answer is negative: we provide a class where learning to sample is easy even if online learning is computationally intractable under standard cryptographic assumptions.  

\begin{theorem}
\label{thm:separation}
    {There is a class $C$ of $n$-variate Boolean functions such that the following hold:
    \begin{enumerate}
        \item (Online Learning is Hard) No {polynomial in $n$ time} algorithm can online learn $C$ in the mistake bound model assuming that one-way functions exist.
        
        \item (Learning to Sample is Easy) PAC learning to sample $C$ can be done in polynomial in $n$ time and sample complexity.
    \end{enumerate}}
\end{theorem}

The concept class is a modification of the class in \cite{blum1994separating}, where they provide a computational separation between offline and online PAC learning. More recently it was used to show computational separations between online and private/replicable PAC learning \cite{bun2020computational,kalavasis2024computational}.

The concept class in \cite{blum1994separating} is based on the classical construction of pseudorandom functions from pseudorandom generators by \cite{GGM-PRF}, described as follows. 

{For ease of notation, we will use $\{0, 1\}^n$ instead of $\{-1, 1\}^n$ as the hypercube in this section. }  {We start with the notion of pseudorandom generators and pseudorandom functions.

\begin{definition}[Pseudorandom Generator]
Let $G$ be a polynomial-time computable function such that for any $n$ and any $s \in \{0, 1\}^n$, $G(s) \in \{0, 1\}^{\ell(n)}$, where $\ell(n)$ is a polynomial in $n$. We say $G$ is a pseudorandom generator if 
\begin{enumerate}
    \item $\ell(n) > n$ for all $n$, and 
    \item for any polynomial-time algorithm $A$, there is a function $\negl$ such that \[\left| \Pr_{s \sim \unif(\{0, 1\}^n)} \left[ A(G(s))=1 \right]- \Pr_{r \sim \unif(\{0, 1\}^{\ell(n)})} \left[ A(r)=1 \right] \right| \le \negl(n), \]
    and $\negl(n) \lesssim 1/n^c$ for any $c > 0$. 
\end{enumerate}
\end{definition}
\begin{definition}[Pseudorandom Function Family]
Let $\{f_s\}_{s \in \{0, 1\}^n}$ be a family of functions such that every $f_s:\{0, 1\}^n \to \{0, 1\}^n$ is polynomial-time computable given $s$. We say $\{f_s\}_s$ is a pseudorandom function family, if for any polynomial-time algorithm $A$, there is a function $\negl$ such that \[\left| \Pr_{s \sim \unif(\{0, 1\}^n)}[A^{f_k(\cdot)}(1^n)=1] - \Pr_{f \sim \unif(\cF_n)}[A^{f(\cdot)}(1^n)=1]\right| \le \negl(n), \] where $\negl(n) \lesssim 1/n^c$ for any $c > 0$, and $\cF_n= \{f:\{0, 1\}^n \to \{0, 1\}^n\}$ . Here $A^{f(\cdot)}(1^n)$ means the algorithm $A$ has query access to $f$, and runs in time polynomial in the input length, which is $n$. 
\end{definition}
}

We will use the following pseudorandom function family from \citet{GGM-PRF}, {which is implementable under the assumption that one-way functions exist}. 
\begin{theorem}[\cite{GGM-PRF}]
\label{thm:PRG-PRF}
Let $G: \{0, 1\}^k \to \{0, 1\}^{2k}$ be a pseudorandom generator. Define $G_0(s)$ and $G_1(s)$ as the first and last $k$ bits of $G(s) \in \{0, 1\}^{2k}$. For any $s, x \in \{0, 1\}^k$, define \[f_s(x) = G_{x_1x_2\cdots x_k}(s) := G_{x_k}(G_{x_{k-1}}(\cdots (G_{x_2}(G_{x_1}(s)))). \]
Then $\{f_s\}_{s \in \{0, 1\}^k}$ is a pseudorandom function family. 
\end{theorem}

We will now describe the concept class associated with this family.
For any $k$, given the pseudorandom generator $G : \{0, 1\}^k \to \{0, 1\}^{2k}$ (as well as $G_0, G_1$) in \cref{thm:PRG-PRF}, \citet{blum1994separating} constructs the concept class $\{c_s\}_{s \in \{0, 1\}^k}$, where $c_s:\{0, 1\}^n \to \{0, 1\}$, $n = k(k+1)$. 
For $i \in \{0, 1, \dots, 2^k-1\}$, viewing {index $i$} as a bit string of length $k$, let \[x_s^i = i \circ G'_{i_1}(s) \circ G'_{i_2}(G_{i_1}(s))\circ G'_{i_3}( G_{i_1i_2}(s))\circ\dots\circ G'_{i_k}(G_{i_1i_2\cdots i_{k-1}}(s)) \circ 0^*,\]
where $\circ$ denotes concatenation, $G'_0(s) = G_1(s)$, $G'_1(s) = \lambda$ the empty string, and $x_s^i$ is padded with $0$ if the length is less than $n = k(k+1)$.
Define $c_s(x_s^i) = \lsb(f_s(i))$, where $\lsb(t)$ denotes the least significant (rightmost) bit of $t$, and $c_s(x) = 0$ for any other $x \in \{0, 1\}^n$. 

\begin{example}
    As an example of the above construction, let $k=2$ and consider the toy example where the PRG is the mapping
\begin{table}[!ht]
    \centering
    \begin{tabular}{c|c|c|c}
        $u$ & $G(u)$ & $G_0(u)$ & $G_1(u)$ \\
        \hline
        $00$ & $1000$ & $10$ & $00$ \\
        $01$ & $0110$ & $01$ & $10$ \\
        $10$ & $0111$ & $01$ & $11$ \\
        $11$ & $1101$ & $11$ & $01$
    \end{tabular}
\end{table}

    Then let us pick $s= 10$ and $c_s : \{0,1\}^6 \to \{0,1\}$. We now fix $i \in \{0,1,2,3\}$ with $i = i_1 i_2$ in binary. Then $f_s(i) = G_{i_2} (G_{i_1}(s))$. Now we have to compute $x_s^i$ which should have length $n=6$, where
    \[
    x_s^{i_1i_2} = i_1i_2 \circ G_{i_1}'(s)
    \circ G_{i_2}'(G_{i_1}(s)) \circ 0^*
    \]
    For example, when $i = 00$, then $G_0'(10) = G_1(10) = 11$ and $G'_0(G_0(10)) = G'_0(01) = G_1(01) = 10$ and so $x_s^{00} = 001110$. Also, $f_s(00) = G_0(G_0(10)) = G_0(01)  =01$. Similarly, one gets the following table:
    \begin{table}[!ht]
    \centering
    \begin{tabular}{c|c|c|c}
        $i$ & $x_s^i$ & $f_s(i)$ & $c_s(x_s^i)=\lsb(f_s(i))$ \\
        \hline
        $00$ & $001110$ & $01$ & $1$ \\
        $01$ & $011100$ & $10$ & $0$ \\
        $10$ & $100100$ & $11$ & $1$ \\
        $11$ & $110000$ & $01$ & $1$
    \end{tabular}
\end{table}
\end{example}

The crucial observation is that $x_s^i$ contains all the information needed to compute $x_s^j$ for any $j > i$, which implies the following lemma. 

\begin{lemma}[\cite{blum1994separating}]
There is an algorithm $\textsc{Compute-Forward}$ that, given as input $x_s^i$ and $j > i$, computes $x_s^j$ and $c_s(x_s^j)$ in $\poly(n)$ time. 
\end{lemma}

To see an intuition about that, let us again focus on the $k=2$ case. The construction of the examples $x_s^i$ may be viewed as a binary tree of depth $2$, whose root is $s$, and whose children are obtained by applying $G_0$ (left) and $G_1$ (right). Thus, the leaf corresponding to index $i=i_1 i_2$ is
\[
f_s(i)=G_{i_2}(G_{i_1}(s)).
\]
The important observation is that whenever the path for $i$ goes left, then the special string $x_s^i$ stores the value of the right sibling. To see this, recall that 
\[
\begin{aligned}
x_s^{00} &= 00 \circ G_1(s) \circ G_1(G_0(s)),\\
\end{aligned}
\]
Here, starting from $s$, the path first moves left because $i_1 = 0$ and it stores $G_1(s)$ (the right child value). Then the path again moves left ($i_2 = 0$) and it stores $G_1(G_0(s))$ (the right child of the node $G_0(s)$). In a similar spirit, the other special strings are
\[\begin{aligned}
x_s^{01} &= 01 \circ G_1(s) \circ 0^*,\\
x_s^{10} &= 10 \circ G_1(G_1(s)) \circ 0^*,\\
x_s^{11} &= 11 \circ 0^*.
\end{aligned}
\]
Therefore, if $j>i$, then at the first coordinate where $i$ and $j$ differ, we have $i_t=0$ and $j_t=1$. But this value is already stored in $x_s^i$ (since $i$ moved left at that level of the tree). Given this value, the remaining part of the path to $j$ can be computed using the public functions $G_0$ and $G_1$. Hence, any $x_s^j$ is efficiently computable from $x_s^i$ whenever $j > i$.

The existence of the function $\textsc{Compute-Forward}$ is the key fact to show that $\{c_s\}_s$ is (distribution-free) PAC-learnable in polynomial time. Their learning algorithm is improper, that is, the hypothesis it outputs is in a larger class $H = \{h_{s, x,b}\}$ where $s \in \{0, 1\}^k, x \in \{0, 1\}^n, b \in \{0, 1\}$, and $h_{s, x, b}(y)$ can be described as follows: 
\begin{enumerate}
    \item Let $i$ be the first $k$ bits of $x$ and $x'$ be the rest. And let $j$ be the first $k$ bits of $y$ and $y'$ be the rest. 
    \item If $j < i$, then outputs $0$.
    \item If $j = i$ and $y'=x'$, then outputs $b$. 
    \item If $j > i$, then let $(x_s^j, c_s(x_s^j)) \gets \textsc{Compute-Forward}(x, j)$, and outputs $c_s(x_s^j) \ind[x_s^j = y]$. 
\end{enumerate}

\begin{theorem}[{\cite[Theorem 3.1]{blum1994separating}}]
\label{thm:ows-PAC-learn}
There is an algorithm that PAC learns $\{c_s\}_{s\in \{0, 1\}^k}$ with polynomial sample and time complexity. 
That is, there exists an algorithm that 
for any distribution $D$ over $\{0, 1\}^n$ and $s\in \{0, 1\}^k  $, given $\varepsilon ,\delta > 0$, and i.i.d.\ samples $\{x_i, c_s(x_i)\}_{i \in [m]}$, where $x_i \sim D$ and $m \ge \poly(n, 1/\varepsilon , \log 1/\delta)$, outputs a hypothesis $h\in H$ such that $\Pr_{x \sim D} [c_s(x) \ne h(x)] \le \varepsilon $, with probability at least $1 - \delta$. 
\end{theorem}

After inspecting the algorithm, one of our key observations is the following:
\begin{observation}
    The algorithm of \citet{blum1994separating} only uses the positive samples it has seen. 
\end{observation}

To sketch the proof, the algorithm only collects the positive samples and computes the smallest index (the first $k$ bits) among the positive samples, denoted by $i^*$, then the algorithm outputs the hypothesis $h_{s, x_s^{i^*}, c_s(x_s^{i^*})}$. It is easy to verify that this hypothesis is consistent with all the samples, and thus the algorithm achieves low population error with high probability since the size of $\{h_{s, x, b}\}$ is $2^{O(n)}$. 

Since the PAC learning algorithm only uses positive samples, to achieve PAC learning to sample, one only needs a sampler for $h_{s, x_s^{i^*}, c_s(x_s^{i^*})}$ given $x_s^{i^*}, c_s(x_s^{i^*})$. This is because if we present {$N^+$} i.i.d.\ uniform positive samples to the algorithm, the (distribution of the) resulting hypothesis will be the same as if we were presenting i.i.d.\ labeled samples uniformly over $\{0, 1\}^n$ {containing $N^+$ positive samples, since the algorithm only depends on the positive samples}.  For the sampler, a natural idea is {rejection sampling:} uniformly sample $j \in [i^*, 2^k]$ and compute $(x_s^j, c_s(x_s^j)) \gets \textsc{Compute-Forward}(x_s^{i^*}, j)$ until $c_s(x_s^j) = 1$, then output $x_s^j$. However, there is a issue with this approach. There might not be many $j > i^*$ such that $c_s(x_s^j) = 1$, so the sampling algorithm needs to sample a lot of $j$'s to find one positive sample. It turns out that this is not an issue when we consider \emph{a random choice of key $s$.}

First, we can show that with high probability over $s$, the fraction of $i$'s that $c_s(x_s^i) = 1$ will not be too small, by the pseudorandomness of $f_s$. 

\begin{lemma}
\label[lemma]{lem:s-good}
Consider the pseudorandom family $\{f_s\}$ of \cite{GGM-PRF}.
For any polynomial $p(n)$, \[\Pr_{s \in \{0, 1\}^k} \left[ \Pr_{i \in \{0, 1\}^k} \left[ \lsb(f_s(i))=1 \right] \le \frac{1}{6} \right] < \frac{1}{p(n)}. \]
\end{lemma}
\begin{proof}
For the sake of contradiction, assume there exists a polynomial $p(n)$ such that \[\Pr_{s \in \{0, 1\}^k} \left[ \Pr_{i \in \{0, 1\}^k} \left[ \lsb(f_s(i))=1 \right] \le \frac{1}{6} \right] \ge \frac{1}{p(n)}, \]
Then we can design an algorithm $\cA^{f(\cdot )}$ that distinguishes pseudorandom functions from uniformly random functions as follows:
\begin{enumerate}
    \item Sample $i_1,\dots, i_{n}$ independently and uniformly from $\{0, 1\}^k$. 
    \item Output $\ind \left[ \left| \{i_j: \lsb(f(i_j))=1\} \right| \ge \frac{n}{3} \right]$. 
\end{enumerate}
\paragraph{Case 1: $f$ is pseudorandom} 
Let $E_1$ denote the event that $\Pr_{i \in \{0, 1\}^k} \left[ \lsb(f_s(i))=1 \right] \le \frac{1}{6}$. 
In this case, we have \begin{align*}
    \Pr_{s, \{i_j\}_{j\in [n]}}[\cA^{f_s(\cdot)}(1^n) = 1] \le \Pr_{s, \{i_j\}_{j\in [n]}} [\cA^{f_s(\cdot)}(1^n) = 1 | E_1] \Pr_s[E_1] + (1 - \Pr_s[E_1]). 
\end{align*}
Note that for any $s$ such that $E_1$ happens, by Hoeffding's inequality, \begin{align*}
    \Pr_{\{i_j\}_{j\in [n]}}\left[\left| \{i_j: \lsb(f_s(i_j))=1\} \right| \ge \frac{n}{3}\right] \le e^{-n/18}. 
\end{align*}
Therefore, \begin{align*}
    \Pr_{s, \{i_j\}_{j\in [n]}}[\cA^{f_s(\cdot)}(1^n) = 1] \le e^{-n/18} \Pr_s[E_1] + (1 - \Pr_s[E_1]) \le 1 - \frac{1}{2p(n)}. 
\end{align*}

\paragraph{Case 2: $f$ is uniformly random}
When $i_j$'s are distinct (let $E_2$ denote this event), $\lsb(f(i_j))$ are independent $\text{Bern}(1/2)$ for $j \in [n]$. Thus, \begin{align*}
    \Pr_{f,\{i_j\}_{j\in [n]}}\left[\left| \{i_j: \lsb(f(i_j))=1\} \right| \le \frac{n}{3}\right] &\le \Pr_{f, \{i_j\}_{j\in [n]}}\left[\left| \{i_j: \lsb(f(i_j))=1\} \right| \le \frac{n}{3}\middle| E_2 \right] + \Pr_{\{i_j\}_{j \in [n]} }[\overline{E_2}] \\
    &\le e^{-n/18}  + \Pr[\exists j_1 \ne j_2, i_{j_1} = i_{j_2}] \\
    &\le e^{-n/18}+ n^2 2^{-k} \\
    &\le \frac{1}{4p(n)}. 
\end{align*}
That is, $
\Pr_{f,\{i_j\}_{j\in [n]}}[\cA^{f(\cdot)}(1^n)=1] \ge 1 - \frac{1}{4 p(n)}.$

We have showed that \begin{align*}
\Pr_{f,\{i_j\}_{j\in [n]}}[\cA^{f(\cdot)}(1^n)=1] - \Pr_{s, \{i_j\}_{j\in [n]}}[\cA^{f_s(\cdot)}(1^n) = 1] \ge \frac{1}{4 p(n)},
\end{align*}
which contradicts the pseudo-randomness of $\{f_s\}_s$.
\end{proof}

Let $S_\good=\{s: \Pr_{i \in \{0, 1\}^k} \left[ \lsb(f_s(i))=1 \right] \ge \frac{1}{6}\}$, then the aforementioned {rejection} sampling algorithm is efficient for $h_{s, x_s^{i^*}, c_s(x_s^{i^*})}$ for $s \in S_\good$. Therefore, we can consider the subclass $\{c_s\}_{s \in S_\good}$, which is also PAC learnable. 

\begin{theorem}
\label{thm:PAC-learn-to-sample-OWF}
There exists an algorithm that PAC learns to sample $\{c_s\}_{s \in S_\good}$ with polynomial sample and time complexity.
That is, for any $s^* \in S_\good$, 
there exists an algorithm that given $\delta, \varepsilon >0$ and $m = \frac{1}{\varepsilon }\log \frac{1}{\delta}$ i.i.d.\ samples from $U^+_{c_{s^*}}$, outputs $x \in \{0, 1\}^n$ such that $\tv{\law(x)}{U_{c_s^*}^+} \le \varepsilon $, with probability at least $1 - \delta$. 
\end{theorem}
\begin{proof}

Recall that the PAC learning algorithm in \cref{thm:ows-PAC-learn} only collects the positive samples and outputs a hypothesis $h$ based on the smallest index $i^*$ among the positive samples $x_{s^*}^{i}$. Note that the hypothesis only makes mistakes on $x_{s^*}^i$ where $i < i^*$ and $c_{s^*}(x_{s^*}^i)=1$. Let $M=|\{i: c_{s^*}(x_{s^*}^i)=1\}|$ and $M' = |\{i < i^*:c_{s^*}(x_{s^*}^i)=1\}|$. Then $\tv{U^+_{h}}{U^+_{c_s^*}} = \frac{M'}{M}$. Given i.i.d.\ samples $x_{s^*}^{i_1},\dots, x_{s^*}^{i_m}$ from $U^+_{c_{s^*}}$, we have \[\Pr[M' \ge \varepsilon M] =\Pr[i^* > i^{(\varepsilon M)}]=\Pr[\forall j\in [m], i_j > i^{(\varepsilon M)}] = (1 - \varepsilon )^m \le \delta,\]
where $i^{(\varepsilon M)}$ denote the $(\varepsilon M)$-th smallest index among $\{i:c_{s^*}(x_{s^*}^i)=1\}$. Therefore with probability at least $1-\delta$, $\tv{U^+_{h}}{U^+_{c_s^*}} \le \varepsilon $. 

It remains to show that $U_h^+$ can be exactly sampled efficiently. Indeed for $s^* \in S_\good$ and hypothesis $h = h_{s, x_s^{i^*}, c_s(x_s^{i^*})}$, one can independently uniformly sample $i \in \{i^*, i^*+1,\dots, 2^k\}$ until $c_{s^*}(x_{s^*}^i)  =1$. This is efficient because \[\Pr_{i \in \{i^*, i^*+1,\dots, 2^k\}}[c_{s^*}(x_{s^*}^i)=1] = \frac{M-M'}{2^k - i^*} \ge \frac{(1-\varepsilon )M}{2^k} \ge \frac{1-\varepsilon }{6}, \] which means the sampling algorithm only needs to draw constantly many independent $i$ in expectation. 
\end{proof}

{ Moreover, one can show as a direct extension of \cite[Theorem~3.2]{blum1994separating} that no algorithm can online-learn larger than $2/3$ fraction of the concepts in $\{c_s\}_{s \in \{0, 1\}^k}$ in the mistake-bound model \cite{blum1994separating} or in the no-regret model \cite{bun2020computational}. For simplicity, we consider the mistake-bound model. 
\begin{theorem}
\label{thm:no-online-for-const-frac}
For any polynomial-time online learning algorithm $\cA_{\onl}$ and any polynomial $p(n)$, \[\Pr_{s \sim \unif(\{0, 1\}^k)}[\mathrm{mistakes}(\cA_{\onl}, c_s) \le p(n)] \le 2/3,\]
where $\mathrm{mistakes}(\cA_{\onl}, c_s)$ is the number of mistakes made by $\cA_{\onl}$ given any order of samples in $\{0, 1\}^n$ labeled by $c_s$. 
\end{theorem}
\begin{corollary}
\label{cor:no-online-for-good}
Assume one-way functions exist. Then no polynomial-time algorithm can online-learn $\{c_s\}_{s\in S_{\good}}$.
\end{corollary}

\begin{proof}
{Since the existence of one-way functions implies the existence of pseudorandom generators, one can construct the concept class $\{c_s\}_{s \in S_\good}$ discussed in this section.}
By \cref{lem:s-good}, $\Pr_s[s\in S_{\good}] \ge 1 - \frac{1}{p(n)}$ for any polynomial $p(n)$. Then any online learner $L$ for $\{c_s\}_{s\in S_{\good}}$ would contradict \cref{thm:no-online-for-const-frac}.
\end{proof}

{Hence, this gives us the desired computational separation: the class of concepts $\{c_s\}_{s \in S_{\good}}$ is efficiently PAC learnable to sample but, under the existence of one-way functions, there is no polynomial-time online learner.}

To show \cref{thm:no-online-for-const-frac}, we will need the following lemma from \cite{blum1994separating}. 
\begin{lemma}[{\cite[Corollary~3.4]{blum1994separating}}]
\label[lemma]{lem:blum-cor}
For any polynomial-time algorithm $\cA$ and polynomial $q$, for sufficiently large $k$, for all $i\in \{0, 1\}^k$, \[\left| \Pr_{s \sim \unif(\{0, 1\}^k)}[\cA^{\cO} (x_s^i, c_s(x_s^i))=1] - \Pr_{s \sim \unif(\{0, 1\}^k)}[\cA^{\cO} (x_s^i, 1-c_s(x_s^i))=1] \right| < \frac{1}{q(k)}, \]
where $\cO$ is an oracle that on any input $j > i$ outputs $(x_s^j, c_s(x_s^j))$. 
\end{lemma}
\begin{proof}[Proof of \cref{thm:no-online-for-const-frac}]
Suppose there exist a polynomial-time learning algorithm $\cA_{\onl}$ such that \[\Pr_{s \sim \unif(\{0, 1\}^k)}[\text{mistakes}(\cA_{\onl}, c_s) \le p(n)] > 2/3\] for some polynomial $p(n)$. Then consider an adversary that presents samples in reverse order $x_s^{2^k}, x_s^{2^k-1},\dots$. Then $\cA_{\onl}$ makes a mistake on at most $\frac15$ of the first $5p(n)$ samples presented, for more than $2/3$ fraction of $s \in \{0, 1\}^k$. Therefore, there exists a ${t} \in [2^k - 5p(n)+1,2^k]$ such that \begin{align*}
    \Pr_{s \sim \unif(\{0, 1\}^k)}[\cA_{\onl} \text{ mistakes } x_s^{t}] < \frac1{5p(n)}\left( \frac{2}{3}\cdot p(n) + \frac{1}{3}\cdot 5p(n) \right) = \frac{7}{15}. 
\end{align*}
Thus, one can construct from $\cA_{\onl}$ an algorithm $\cA$ that contradicts \cref{lem:blum-cor}. On input $(x_s^t, b)$, the algorithm $\cA$ uses the oracle to compute $(x_s^j, c_s(x_s^j))$ for $j > i$ to simulate the samples for online learner $\cA_{\onl}$ from the adversary. Then $\cA$ outputs $1$ iff the $\cA_{\onl}$ outputs $b$ on $x_s^t$. We will have \begin{align*}
     \Pr_{s \sim \unif(\{0, 1\}^k)}[\cA^{\cO} (x_s^i, c_s(x_s^i))=1] - \Pr_{s \sim \unif(\{0, 1\}^k)}[\cA^{\cO} (x_s^i, 1-c_s(x_s^i))=1] > \frac{8}{15} - \frac{7}{15} = \frac{1}{15},
\end{align*}
which contradicts \cref{lem:blum-cor}. 
\end{proof}
}

\subsection{PAC Learning to Sample with ERM Oracles}
\label{sec:new-reduction}
Inspired by the previous computational separation, we establish a collection of reductions from offline PAC learning (instead of online) to PAC learning to sample. To state our result we will need some background on the oracles we are going to use.

\subsubsection{Background}

We are going to replace the online learner with the offline learner. In particular, we will use offline Empirical Risk Minimization (ERM) defined as follows:
\begin{definition}
    [ERM Oracle]
    \label{def:erm}
    Let $C$ be a class of $n$-variate Boolean concepts and let $S$ be a set of $m$ labeled examples $(x_i,y_i)$ for $i \in [m]$. The ERM oracle $\cA_{\erm}^C$ for $C$ gets as input $S$ and returns a hypothesis in the set $\argmin_{c \in C} \sum_{i \in [m]} \ind\{c(x_i) \neq y_i\}$ in time $T_{\erm}(n, m)$. 
    We say that the ERM oracle is efficient if $T_{\erm}(n, m) = \poly(n, m)$. 
\end{definition}

For our results we will also need the following extension of the sampling and counting oracles of \citet{de2014learning}.

\begin{definition}[Approximate Counting for Intersections]
Let $C$ be a concept class of $n$-variate Boolean functions. A randomized algorithm $\cA^{\bigcap C}_{\cnt}$ is an approximate counting algorithm for intersections of $C$ if given any $k \ge 1$, $f_1,\dots, f_k \in C$ and $ \varepsilon,\delta > 0$, it outputs a value $\wh  \alpha$ in time $T_{\cnt}^{\bigcap C}(n,k,1/\varepsilon ,1/\delta)$ such that with probability at least $1 - \delta$, \[\frac{1}{1+ \varepsilon } \Pr_{x\sim \unifcube}[\forall i\in [k],\,f_i(x) = 1] \le \wh  \alpha \le (1+ \varepsilon) \Pr_{x\sim \unifcube}[\forall i\in [k],\,f_i(x) = 1]. \]
We say that the approximate counting algorithm for intersections is efficient if \[
T_{\cnt}^{\bigcap C}(n,k,1/\varepsilon ,1/\delta) = \poly(n, k, 1/ \varepsilon , \log(1/ \delta)).
\]
\end{definition}

\begin{definition}[Approximate Uniform Generation for Intersections]
Let $C$ be a concept class of $n$-variate Boolean functions. A randomized algorithm $\cA^{\bigcap C}_{\gen}$ is an approximate uniform generation algorithm for intersections of $C$ if for any $k \ge 1$, $f_1,\dots, f_k \in C$ and $\delta > 0$, there is a distribution $D$ supported on $\bigcap_{i=1}^k f_i^{-1}(1)$ such that for each $x \in \bigcap_{i=1}^k f_i^{-1}(1)$,
\[
\frac{1}{1+\delta} \cdot \frac{1}{|\bigcap_{i=1}^k f_i^{-1}(1)|} \le D(x) \le (1+\delta) \cdot \frac{1}{|\bigcap_{i=1}^k f_i^{-1}(1)|},
\]
and, given $k$, $f_1,\dots, f_k$ and $\delta$, the algorithm runs in time $T_{\gen}^{\bigcap C} (n,k ,1/\delta)$ and either outputs a point $x \in \bigcap_{i=1}^k f_i^{-1}(1)$ distributed according to $D$, or reports failure.
Moreover, the probability of failure is at most $\delta/2$.
We say that the approximate uniform generation algorithm for intersections is efficient if $T_{\gen}^{\bigcap C}(n, k, 1/\delta) = \poly(n ,k, 1/ \delta)$.
\end{definition}

{The reader should in general think of the number of intersections $k$ as $O(n)$; our results will be explicit on that. When $k=1$, we recover the definition of approximate counting and sampling oracles for a concept class $C$, as defined in \Cref{sec:resultDDS}.
Since $\big|D(x) - 1/|\bigcap_{i=1}^k f_i^{-1}(1)|\big| \le \delta/|\bigcap_{i=1}^k f_i^{-1}(1)|$ for each $x \in \bigcap_{i=1}^k f_i^{-1}(1)$, we have $\tv{D}{\unif(\bigcap_{i=1}^k f_i^{-1}(1))} \le \delta/2$.
Hence, if we replace a failure by an arbitrary fixed point of $\{\pm 1\}^n$, the law of the output is within total variation distance $\delta$ of $\unif(\bigcap_{i=1}^k f_i^{-1}(1))$, since a failure happens with probability at most $\delta/2$.
We use this convention in \cref{sec:densifier,sec:tradeoff}, and this total variation bound is the only property of the approximate uniform generation algorithm used there.}

\subsubsection{Intermediate Key Result}
\label{sec:densifier}
Our main result for this section is the following algorithm that uses the ERM oracle together with the approximate sampling/counting oracles for intersections of $C$ to get a sampler that achieves large breadth and non-trivial hallucination rate. 
\begin{theorem}
    \label{thm:densifier-ERM}
    Let $C$ be a concept class of $n$-variate Boolean concepts with induced class of languages $\cL$. 
    Then there is an algorithm that given access to an offline ERM oracle for $C$, 
    an approximate uniform generation oracle for intersections of $C$, and an approximate counting oracle for intersections of $C$,
    gets as input
     $\eps,\delta \in (0,1)$, i.i.d.\ samples from $U^+_{L^\star}$ with any $L^\star \in \cL$ and $\wh \alpha \in [2^{-n},1]$. If $\alpha \le \wh \alpha\le (1 + \varepsilon )\alpha$, where $\alpha = \Pr_{x \sim \unifcube}[L^\star(x)=1]$, the algorithm outputs a hypothesis $h$ with
    \begin{enumerate}
    \item $\Pr_{x \sim U_{L^\star}^+} [h(x) = 1] \ge 1-\varepsilon$;
    \item $\Pr_{x\sim \unifcube}[h(x) = 1] \le \alpha/\gamma$, where $\gamma = \delta/(128 n \log \frac{8n|C|}{\delta})$,
\end{enumerate}
with probability at least $1-\delta$. 
Moreover, the algorithm \begin{enumerate}
    \item uses $ m = O(\frac{n}{\varepsilon } \log\frac{|C|}{\delta})$ i.i.d. samples from the true target distribution $U^+_{L^\star}$,
    \item makes at most $n$ queries to the ERM oracle with sample size $O(m)$, and runtime $T_{\erm}(n, O(m))$,
    \item makes at most $n$ queries to the approximate counting oracle for intersections of $k \le n$ concepts in $C$ with accuracy $\eps$, confidence $\delta'=\Theta(\delta/n)$, and runtime $T_{\cnt}^{\bigcap C}(n,k,1/\varepsilon ,1/\delta')$,
    \item makes at most $n$ queries to the approximate uniform generation oracles for intersections of $k \le n$ concepts in $C$ with accuracy $\delta''=\Theta({\delta}/({n \log \frac{n|C|}{\delta}}))$, and runtime $T_{\gen}^{\bigcap C}(n, k, 1/\delta'')$.
\end{enumerate}
The algorithm runs in time \[\poly(n, \log|C|,1/\epsilon,\log(1/\delta),T_{\erm}(n, O(m)), T_{\cnt}^{\bigcap C}(n,k,1/\varepsilon ,1/\delta'),T_{\gen}^{\bigcap C}(n, k, 1/\delta'')).\]
\end{theorem}
Inspecting the properties of the output $h$, one can observe that covers essentially most of $L^\star$ (large breadth) and it achieves a hallucination guarantee of the form:
\[
\sum_{x \in \{-1,1\}^n} \ind\{h(x) = 1\}
\lesssim 
(1/\gamma) \cdot \sum_{x \in \{-1,1\}^n} \ind\{x \in L^\star\}
\]
Intuitively, this means that $h$ hallucinates by a factor at most $1/\gamma = O(n\log\frac{n|C|}\delta/\delta)$ compared to the mass of the true language. In other words, if $\alpha$ is the fraction of Boolean strings in $L^\star$, the language induced by $h$ contains at most $\alpha /\gamma$ many strings.

\begin{remark}
Given the guarantee of our algorithm it is not clear how one can turn these guarantees to {PAC learning to sample guarantees} (i.e., how we can compress this $1/\gamma$ factor, {and how to accurately estimate $\alpha$, the mass of the true language}). 
A similar phenomenon appears in the work of \cite{de2014learning}. In fact,
the guarantees that we get in the above intermediate result correspond to the problem of learning a \emph{densifier} for the true language, as introduced in \citet{de2014learning}. The important observation is that the above algorithm does not make use of the SQ learner. In the next section, {we will show the PAC learning to sample guarantees by our densifier, the SQ learner, and enumerating the guesses for the true mass, following \cite{de2014learning}. }
\end{remark}

\begin{algorithm}[H]
\caption{Algorithm for \Cref{thm:densifier-ERM}}
\label{alg:den}
\SetKwInput{Input}{Input}
\SetKwInput{Oracle}{Oracle Access}
\SetKwInput{Output}{Output}
\Input{Sampler access to $U_{L^\star}^+$, parameters $\varepsilon ,\delta > 0$, a value $\wh  \alpha \in [2^{-n}, 1]$}
\Oracle{ERM Oracle $\cA_{\rm ERM}^C$, Sampling Oracle $ \cA_{\gen}^{\bigcap C}$, Counting Oracle $ \cA_{\cnt}^{\bigcap C}$}
Draw $N_+ = \frac{n+1}{\varepsilon }\log \frac{4|C|}{\delta}$ i.i.d.\ samples $x_1^+,\dots, x_{N_+}^+$ from $U_{L^\star}^+$\;
$h^0 \gets \cA_{\erm}^C(\{(x_j^+,1)\}_{j\in [N^+]})$\;
\For{$i \gets 1$ \KwTo $n$}{
    Draw $N_- = 2 \log \frac{8n|C|}{\delta}$ i.i.d.\ samples $x_1^-,\dots, x_{N_-}^-$ from $\cA^{\bigcap C}_{\gen}(i, \{h^0,\dots, h^{i-1}\},\delta/(8nN^-))$\;
    $h^i \gets \cA_{\erm}^C(\{(x_j^+,1)\}_{j\in [N^+]} \cup \{(x_j^-,-1)\}_{j\in [N^-]})$\;
    $\wh  \alpha_i \gets \cA^{\bigcap C}_{\cnt}(i+1, \{h^0,\dots, h^{i}\}, \varepsilon ,\delta/(4n))$\;
    \If(\tcp*[f]{$\gamma = \delta/(64 n N^-)$}){$\wh  \alpha_i \le \wh  \alpha/(\gamma (1+ \varepsilon )^2)$}{
        \Return{$h=\bigcap_{j=0}^i h^j$}\;
    }
}
\end{algorithm}

Before proving the above guarantees, we will need the following lemmas, that argue about how to control the breadth and hallucination of the set of consistent hypotheses. 

\Cref{lem:breadth} guarantees that \emph{any} hypothesis in $C$ that is consistent with a sufficient large set of positive examples drawn from the true target distribution $U_{L^\star}^+$ achieves non-trivial expressivity, i.e., $\Pr_{x \sim U_{L^\star}^+}[f(x) = 1] \ge 1 - \varepsilon$.

\begin{lemma}
[Breadth of Conistent Hypotheses]
\label[lemma]{lem:breadth}
For any $\varepsilon ,\delta > 0$, if $N_+ \ge \frac{1}{\varepsilon }\log \frac{|C|}{\delta}$, 
then with probability at least $1 -\delta$ over the draw of $x_1^+, \dots, x_{N^+}^+ \overset{\rm i.i.d.}{\sim} U_{L^\star}^+$, 
any function $f\in C$ such that $f(x_i^+)=1$ for all $i \in [N^+]$ 
satisfies $\Pr_{x \sim U_{L^\star}^+}[f(x) = 1] \ge 1 - \varepsilon $. 
\end{lemma}
\begin{proof}
By the union bound, \begin{align*}
    &\Pr_{x_1^+,\dots, x_{N^+}^+ \overset{\rm i.i.d.}{\sim} U_{L^\star}^+ }\left[\exists f \in C, f(x_i^+) = 1, i \in [N^+], \Pr_{x\sim U_{L^\star}^+}[f(x) = 1] < 1 - \varepsilon  \right] \\
    &\le |C| \Pr_{x_1^+,\dots, x_{N^+}^+ \overset{\iid}{\sim} U_{L^\star}^+ }\left[f(x_i^+) = 1, i \in [N^+], \Pr_{x\sim U_{L^\star}^+}[f(x) = 1] < 1 - \varepsilon\right] \\
    &< |C| (1-\varepsilon )^{N^+} \\
    &\le \delta. 
\end{align*}
\end{proof}

Regarding hallucinations, we are now ready to present our main idea which is a \emph{volume contraction} argument that works as follows: After the application of \Cref{lem:breadth}, we obtain some hypothesis $h^0$ that achieves breadth but may have extreme hallucinations. However, because its mass is much larger compared to the true language, we can use the approximate sampling oracle for $h^0$ to generate few random examples. Crucially, because of the gap between the mass of $h^0$ and the true language, we can, with high probability, consider that all these examples are \emph{negative} for the true hypothesis. This means that we can employ our offline PAC learner to make progress and get a new consistent hypothesis $h^1$ that also achieves breadth but its hallucinations are reduced.

Next, we iterate the same argument but now generate samples from the intersection of $h^0$ and $h^1$. In particular, at step $i \in [n]$, our lemma shows the following \emph{contraction} property on the volume of our prediction compared to the true hypothesis: 
\begin{lemma}
[Hallucination Contraction]
\label[lemma]{lem:vol-reduce}
{Fix $i \in [n]$.}
For any $c ,\delta > 0$, if $N^- \ge \frac{1}{c} \log \frac{2|C|}{\delta}$, then with probability at least $1 - \delta$ over the draw of $x_1^-,\dots, x_{N^-}^- \overset{\iid}{\sim} \cA^{\bigcap C}_{\rm gen} (i, h^0,\dots, h^{i-1},\delta/2N_-) $, any function $f \in C $ such that $f(x_i^-) = -1$ for all $i \in [N^-]$ satisfies \[
\Pr_{x \sim \unifcube}\left[\left(\bigcap_{j=0}^{i-1} h^j \cap f\right)(x) = 1\right] \le c \cdot  \Pr_{x \sim \unifcube}\left[\left(\bigcap_{j=0}^{i-1} h^j\right)(x) = 1\right].
\]
\end{lemma}

\begin{proof}
Let $g = \bigcap_{j=0}^{i-1} h^j$.
Since $\cA^{\bigcap C}_{\rm gen}$ is an approximate uniform generation algorithm for intersections of $C$, 
we have that $\tv{\law(x_i^-)}{\unif (g^{-1}(1))} \le \frac{\delta}{2N^-}$. Since $x_i^-$ are independent, \begin{align*}
    \tv{\law(x_1^-,\dots, x_{N^-}^-)}{\left( \unif (g^{-1}(1)) \right)^{\otimes N^-}} \le N^- \cdot \frac{\delta}{2N^-} = \frac{\delta}{2}. 
\end{align*}
Following the same argument as in the proof of \cref{lem:breadth}, we have \begin{align*}
    &\Pr_{x_1',\dots, x_{N^-}' \overset{\iid}{\sim} \unif(g^{-1}(1))}\left[\exists f\in C, f(x_i')=-1, i\in [N^-], \Pr_{x\sim \unif(g^{-1}(1))}[f(x)=-1] \le 1- c \right] \\
    &\le |C| (1 - c )^{N^-} \le \delta/2. 
\end{align*}
Moreover, \begin{align*}
    &\Pr_{x_1^-,\dots, x_{N^-}^- \overset{\iid}{\sim}{\cA^{\bigcap C}_{\rm gen}}}\left[\exists f\in C, f(x_i^-)=-1, i\in [N^-], \Pr_{x\sim \unif(g^{-1}(1))}[f(x)=-1] \le 1- c \right] \\
    &\le \Pr_{x_1',\dots, x_{N^-}' \overset{\iid}{\sim} \unif(g^{-1}(1))}\left[\exists f\in C, f(x_i')=-1, i\in [N^-], \Pr_{x\sim \unif(g^{-1}(1))}[f(x)=-1] \le 1- c \right] \\
    &\qquad + \tv{\law(x_1^-,\dots, x_{N^-}^-)}{\left( \unif (g^{-1}(1)) \right)^{\otimes N^-}} \\
    &\le \frac{\delta}{2} + \frac{\delta}{2} = \delta. 
\end{align*}
Thus, with probability at least $1-\delta$, any $f\in C$ that $f(x_i^-) =-1$ for all $i\in [N^-]$ satisfies $\Pr_{x\sim \unif(g^{-1}(1))} [f(x)=1] \le c$, i.e., \begin{align*}
    \frac{\Pr_{x\sim \unifcube} [f(x) = 1, g(x) =1]}{\Pr_{x\sim \unifcube}[g(x)=1]} &= \Pr_{x\sim \unifcube}[f(x)=1|g(x)=1] \\
    &= \Pr_{x \sim \unif(g^{-1}(1))}[f(x)=1] \le c. 
\end{align*}
\end{proof}

We are now ready to prove the main result of this section. 

\begin{proof}[Proof of \cref{thm:densifier-ERM}]
First, we can show that for each iteration $i$ in the for-loop, all the $N_-$ samples $x_1^-,\dots, x_{N^-}^-$ (associated with round $i)$ are labeled as $-1$ by $L^\star$ with high probability. 
Let $g_{i-1} = \bigcap_{j=0}^{i-1} h^j$ and 
$\alpha_{i-1} = \Pr_{x \sim \unifcube}\left[ g_{i-1}(x) = 1 \right]$. 
Since in the $i$-th iteration, it holds that $\wh  \alpha_{i-1} > \wh  \alpha / (\gamma (1+ \varepsilon )^2)$, we have $\alpha_{i-1} \ge \wh  \alpha_{i-1}/(1 + \varepsilon) > \wh  \alpha / (\gamma (1 + \varepsilon )^3) \ge \alpha / (\gamma (1 + \varepsilon )^3) \ge \alpha/ (8 \gamma) $, with probability at least $1 - \delta/(4n)$. 
From the proof of \cref{lem:vol-reduce}, we know that $\tv{\law(x_1^-,\dots, x_{N^-}^-)}{\left( \unif(g_{i-1}^{-1}(1)) \right)^{\otimes N^-}} \le \delta/(8n)$. 
Therefore, for a fixed round $i:$
\begin{align*}
    \Pr_{x_1^-,\dots, x_{N^-}^- \overset{\iid}{\sim}{\cA^{\bigcap C}_{\rm gen}}} [\forall j \in [N^-],L^\star(x_j^-) = -1] &\ge \Pr_{x_1',\dots, x_{N^-}' \overset{\iid}{\sim}\unif(g_{i-1}^{-1}(1))} [\forall j \in [N^-],L^\star(x_j') = -1] - \frac{\delta}{8n} \\
    &\ge \left( 1 - \Pr_{x \sim \unif(g_{i-1}^{-1}(1))}[L^\star(x)=1] \right)^{N^-} - \frac{\delta}{8n} \\
    &\ge \left( 1 - \frac{\Pr_{x\sim \unifcube}[L^\star(x)=1]}{\Pr_{x\sim \unifcube}[g_{i-1}(x)=1]} \right)^{N^-} - \frac{\delta}{8n} \\
    &= \left( 1 - \frac{\alpha}{\alpha_{i-1}} \right)^{N^-} - \frac{\delta}{8n} \\
    &\ge 1 - 8N^-\gamma-\frac{\delta}{8n} = 1 - \frac{\delta}{4n}. 
\end{align*}
As a result, for each $i\in [n]$, with probability at least $1 - \delta/(4n)$, $h^i = \argmin_{f\in C} \sum_{j=1}^{N_+} \ind[f(x_j^+) \ne 1] + \sum_{j=1}^{N_-} \ind[f(x_j^-) \ne -1]$ satisfies $h^i(x_j^+) = 1$, $j\in [N^+]$ and $h^i(x_j^-) = -1$, $j\in [N^-]$, since $L^\star$ achieves this minimum. 
By \cref{lem:breadth}, with probability at least $1 - \delta/4$, any function $f \in C$ such that $f(x_i^+) = 1$ for all $i\in [N^+]$ satisfies $\Pr_{x \sim U_{L^\star}^+}[f(x) = 1] \ge 1 - \varepsilon /(n+1)$. Also, by \cref{lem:vol-reduce}, for each $i \in [n]$, with probability at least $1 - \delta/(4n)$, any function $f \in C$ such that $f(x_i^-) = -1$ for all $i\in [N^-]$ satisfies $\Pr_{x \sim \unifcube}\left[(\bigcap_{j=0}^{i-1} h^j \cap f)(x) = 1\right] \le \frac12 \Pr_{x \sim \unifcube}\left[(\bigcap_{j=0}^{i-1} h^j)(x) = 1\right]$. 

By a union bound, with probability at least $1-\delta$, all the above events hold. In this case, we have that for each $i \in [n]$, \begin{align*}
    \Pr_{x \sim \unifcube}\left[\left(\bigcap_{j=0}^{i} h^j\right)(x) = 1\right] &\le \frac12 \Pr_{x \sim \unifcube}\left[\left(\bigcap_{j=0}^{i-1} h^j\right)(x) = 1\right] \\
    &\le \frac{1}{2^i} \Pr_{x \sim \unifcube} \left[ h^0(x) = 1 \right] \\
    &\le 2^{-i}.
\end{align*}
Thus, $\wh  \alpha_i \le (1 + \varepsilon ) 2^{-i}$. 
Since $\wh  \alpha_n \le (1 + \varepsilon ) 2^{-n} \le \wh  \alpha/(\gamma (1 + \varepsilon )^2)$, there must be an iteration when $i = k\in [n]$, that the algorithm terminates and returns $h = \bigcap_{j=0}^k h^j$ with $\wh  \alpha_k \le \wh  \alpha / (\gamma(1 + \varepsilon )^2)$. 
When this happens, \begin{align*}
    \Pr_{x \sim \unifcube}[h(x) = 1] \le (1 + \varepsilon ) \wh  \alpha_i \le \wh  \alpha /(\gamma (1 + \varepsilon )) \le \alpha/\gamma. 
\end{align*}

Meanwhile, we have $\Pr_{x \sim U_{L^\star}^+}[h^i(x)=1] \ge 1 - \varepsilon /(n+1)$ for each $i = 0, 1, \dots, k$. Then by the union bound, \begin{align*}
    \Pr_{x \sim U_{L^\star}^+}[h(x) = -1] \le (k+1) \cdot \frac{\varepsilon }{n+1} \le \varepsilon . 
\end{align*}
\end{proof}

\subsubsection{Getting TV distance bounds}
\label{sec:sq+guess}
Given a hypotheses $h$ that satisfies the conditions in \cref{thm:densifier-ERM} (a densifier), we can apply the techniques in \cite{de2014learning} as a black box to achieve $\varepsilon  $ TV distance, given an SQ learning algorithm for the class $C$. 

\begin{theorem}
[{\cite[Appendix~C]{de2014learning}}]
\label{thm:sq+guess}
Let $C$ be a concept class of $n$-variate Boolean concepts with induced class of languages $\cL$. Suppose there is an algorithm $\cA^C_{\den}$ that given $\varepsilon ,\delta \in (0, 1)$, i.i.d.\ samples $S$ from $U_{L^\star}^+$ with any $L^\star \in \cL$ and $\wh \alpha \in [2^{-n}, 1]$, if $\alpha \le \wh \alpha \le (1+ \varepsilon )\alpha$, where $\alpha = \Pr_{x \sim \unifcube}[L^\star(x)=1]$, outputs a hypothesis  $h \in C'$ for some concept class $C'$ such that 
\begin{enumerate}
    \item $\Pr_{x \sim U_{L^\star}^+} [h(x) = 1] \ge 1-\varepsilon$;
    \item $\Pr_{x\sim \unifcube}[h(x) = 1] \le \alpha/\gamma$, where $\gamma = \gamma(n,1/\varepsilon ,1/\delta)$,
\end{enumerate}
with probability at least $1-\delta$. 
Then there is an algorithm that, given access to $\cA_{\den}$, an approximate uniform generation and counting oracle for $C'$, and an SQ learning algorithm for $C$, gets as input $\varepsilon ,\delta > 0$, outputs a sampler $G_S$ such that \[\tv{G_S(U)}{U_{L^\star}^+} \le \varepsilon \] with probability at least $1-\delta$. 
Moreover, the algorithm 
\begin{enumerate}
    \item makes $O(n/\varepsilon )$ calls to $\cA_\den$ with accuracy $\varepsilon _\den=\Theta(\varepsilon )$, confidence $\delta_\den=\Theta(\delta)$, and runtime $T_{\den}=T_{\den}(n, 1/\varepsilon _\den, 1/\delta_\den)$,
    \item makes $O(n/\varepsilon )$ calls to the SQ learning algorithm with accuracy $\varepsilon _\sq = \Theta(\varepsilon \gamma)$, confidence $\delta_\sq = \Theta(\delta)$, runtime $t_1=t_1(n, 1/\varepsilon _\sq,1/\delta_\sq)$, and each SQ query takes $t_2=\poly(n)$ time to evaluate with tolerance $\tau=\tau(n,1/\varepsilon _\sq)$. 
    \item uses $m = \poly(t_1, 1/\tau,\log(1/\delta),1/\gamma)$ extra i.i.d.\ samples from the true target distribution $U_{L^\star}^+$,
    \item makes at most $O(n/\varepsilon )$ extra queries to the approximate counting oracle for $C'$ with accuracy $\varepsilon _\cnt = \Theta(\min\{\varepsilon ,\tau\})$, confidence $\delta_\cnt=\Theta(\delta)$, and runtime $T_\cnt=T_{\cnt}(n, 1/\varepsilon _\cnt,1/\delta_\cnt)$, 
    \item makes at most $\poly(n,1/\varepsilon ,t_1,1/\tau,\log(1/\delta),1/\gamma)$ extra queries to the approximate uniform generation oracle for $C'$ with accuracy $\delta_{\gen} = \poly(1/n,\varepsilon ,1/t_1,\tau,\delta,\gamma)$ and runtime $T_\gen=T_\gen(n,1/\delta _\gen)$. 
\end{enumerate}
The algorithm runs in time \[
\poly(n,1/\varepsilon,\log(1/\delta),1/\gamma,T_{\den}, t_1, t_2,\tau^{-1},T_{\cnt},T_\gen). 
\]
{The sampler $G_S$ can be computed in time $O(T_\gen(n, \wt O(\frac1{\delta \varepsilon \gamma}))\frac{1}{\gamma}\log\frac{1}{\delta \varepsilon }). $}
\end{theorem}

There are two main ingredients to \cref{thm:sq+guess} in \cite{de2014learning}: the use of the SQ learning algorithm and the enumeration of the guess $\wh \alpha$ for the true mass $\alpha$. 
\begin{enumerate}
    \item The SQ learning algorithm will run under distribution $U_h^+$ where $h$ is the hypothesis from $\cA_\den$. One can simulate statistical queries using only samples from $U_{L^\star}^+$ and unlabeled samples from $U_{h^+}$ (which can be generated from the uniform generation oracle). Since it is guaranteed that the mass of $h$ is at most $1/\gamma$ times larger than the true mass, we only need to pay an extra $\gamma$ factor on the accuracy parameter of the SQ learning algorithm to get $\varepsilon $ \emph{multiplicative} error, as long as the guess $\wh \alpha$ is accurate. 
    \item We need to enumerate the guess $\wh \alpha = (1+ \varepsilon )^i/2^n$ for $i = 0,1,\dots, O(n/\varepsilon )$, so that one of them in the enumeration satisfies $\alpha \le \wh \alpha \le (1 + \varepsilon )\alpha$. Under each guess, the SQ algorithm gives a hypothesis. Then \cite{de2014learning} runs a hypothesis testing process to select a good concept from the candidates. 
\end{enumerate}

As a result, combining \cref{thm:densifier-ERM} and \cref{thm:sq+guess}, we get the following for PAC learning to sample result. 

\begin{theorem}
\label{thm:PAC-learn-sample-ERM}
Let $C$ be a concept class of $n$-variate Boolean concepts with induced class of languages $\cL$. 
Then there is an algorithm that PAC learns to sample $C$, given access to 
\begin{enumerate}
    \item an efficient offline ERM oracle for $C$,
    \item an efficient approximate uniform generation oracle for intersections of at most $n$ concepts in $C$,
    \item an efficient approximate counting oracle for intersections of at most $n$ concepts in $C$, and 
    \item an efficient SQ learning algorithm for $C$. 
\end{enumerate} 
That is, for any $L^\star \in \cL$, given $\varepsilon ,\delta > 0$ and $m = \poly(n, 1/\varepsilon , 1/\delta, \log|C|)$ i.i.d.\ samples $S$ from $U_{L^\star}^+$, the algorithm outputs a {$\poly(n,1/\varepsilon ,1/\delta, \log|C|)$-time-computable} sampler $G_S$, such that with probability at least $1-\delta$, \[\tv{{G_S(U)}}{U_{L^\star}^+} \le \varepsilon , \]
in $\poly(n, 1/\varepsilon ,1/\delta, \log|C|)$ time. 
\end{theorem}

\subsection{Trade-offs between negative samples and size of intersection}
\label{sec:tradeoff}
In the previous sections, we have seen two examples of PAC learning to sample algorithms:

\begin{itemize}
\item In \cref{sec:ows}, our algorithm requires a PAC learner that only depends on the positive samples and a sampling oracle for the concept class.  

\item In \cref{sec:new-reduction}, our algorithm requires a PAC learner (ERM) that depends on all the samples and a sampling/counting oracle for intersections of the concept class $C$. The intersection could be over at most $n$ hypotheses of $C$. 
\end{itemize}

Therefore, the natural question is whether one could get something in between. That is, the PAC learner has mild dependency on the negative samples, and the intersection would be over $m = o(n)$ hypotheses. 

To formalize the dependency on the negative samples, we can imagine that the training dataset (which contains both positive and negative examples) is being processed by some filter before provided to the learner and the filter keeps all the positive samples and removes some of the negative samples. We make this concrete in the following definition:

\begin{definition}[$p$-Negative-Filtering]
{Fix $0 \leq p \leq 1.$}
Let $S$ be a labeled training set, and $S^+$, $S^-$ be the positive and negative samples in $S$. 
The $p$-negative-filter is defined as $\cF_p(S) = S^+ \cup S_p^-$, 
where each $(x, -1) \in S^-$ is included in $S_p^-$ with probability $p$ independently. 
\end{definition}

{Note that when $p=0$, we only observe positive examples.}
We are interested in PAC learning algorithms that are robust to $p$-negative filters in the following sense:
\begin{definition}
[PAC Learning with $p$-Negative-Filtering]
{Fix $0 \leq p \leq 1.$}
Let $C$ be a concept class of $n$-variate Boolean functions. 
A PAC learner for $C$ with $p$-negative-filtering is an algorithm $\cA_{\pac}^p$ that for any $f\in C$, any distribution $D$ over $\{\pm 1\}^n$, and large enough $N_0$, given any $\delta> 0$ and filtered samples $S_p=\cF_p(\{(x_i, f(x_i))\}_{i\in [N_0]})$ where $x_i \overset{\iid}{\sim} D$, outputs a hypothesis $h \in C'$ {in time $T_{\pac}^p(n,|S_p| ,1/\delta)$} such that $\Pr_{x \sim D}[f(x) \ne h(x)] \le \varepsilon=\varepsilon (n,N_0,1/\delta) $ with probability at least $1 - \delta$ over {$\{x_i\}_{i\in [N_0]}$, $\cF_p$, and the randomness of $\cA_{\pac}^{p}$}, where $\varepsilon (n, N_0,1/\delta)$ is \emph{independent} of $p$ and corresponds to the achievable rate \emph{without} filtering.
We say the PAC learner $\cA_\pac^p$ is efficient if $T_\pac^p(n, |S_p|, 1/\delta) = \poly(n, |S_p|, \log(1/\delta))$. 
\end{definition}
{This means that the learning algorithm does not need all the negative examples, in the sense that even if we remove $1-p$ fraction of the negative examples, the algorithm has the same rate $\varepsilon (n, N_0, 1/\delta)$ as the rate achievable without removing examples (such as $\mathrm{VC}(C)/N_0$ in the realizable PAC learning setting). }
For the above model, a crucial quantity is the amount of positive and negative examples of the true labeling function $f$. We denote the positive fraction by 
\[
    \alpha := \frac{f^{-1}(1)}{2^n}\,.
\]

In our algorithm, we will simulate the filtered samples and use them as the input to the PAC learner. When $p$ and $\alpha$ are super-polynomially small, the PAC learner could possibly only use polynomially many (filtered) samples to achieve superpolynomially small additive error. 
To illustrate this intuition, suppose $p = 0$ and $\alpha = 2^{-\Theta(n)}$. 
Then giving $N$ \iid uniform positive samples to the PAC learner is equivalent to passing in expectation $N/\alpha$ \iid labeled samples uniformly from $\{\pm 1\}^n$ to the filter $\cF_0$, then giving the filtered samples to the learner. That is, setting $N = \poly(n, \log(1/\delta))$ would make the accuracy $\varepsilon $ scale linearly with $\alpha$, which could be translated to a TV distance bound between $U^+_{f}$ and $U^+_{h}$ that is independent of $\alpha$. 

To generalize this to $p > 0$, recall that in \cref{lem:vol-reduce}, the hallucination contraction factor depends on the number of negative samples. For superpolynomially small $p$ and $\alpha$, one could get superpolynomially small contraction factor in each round, using only polynomially many samples. 

The result is as follows.

\begin{theorem}
\label{thm:den-tradeoff}
Let $C$ be a concept class of $n$-variate Boolean concepts with induced class of languages $\cL$.  
Assume that one has access to 
a PAC learner for $C$ (outputting a hypothesis in $C'$) with $p$-negative-filtering, 
an approximate uniform generation oracle for intersections of $C'$, 
and an approximate counting oracle for intersections of $C'$. 
Moreover, assume that for any $N_0 ,\delta_{\pac} > 0$, the PAC learner achieves $\varepsilon_{\pac}(n,N_0,1/\delta_\pac)=R(n,\delta_\pac)/N_0=n^c \poly\log(1/\delta_\pac)/N_0 $ error with probability at least $1-\delta_{\pac}$, where $N_0$ is the number of samples before filtering. 
Then, there is an algorithm that gets as input
 $\eps,\delta \in (0,1)$, i.i.d.\ samples from $U^+_{L^\star}$ with any $L^\star \in \cL$ and $\wh \alpha \in [2^{-n},1]$. If $\alpha \le \wh \alpha\le (1 + \varepsilon )\alpha$, where $\alpha = \Pr_{x \sim \unifcube}[L^\star(x)=1]$, the algorithm outputs a hypothesis $h$ with
\begin{enumerate}
    \item $\Pr_{x \sim U_{L^\star}^+} [h(x) = 1] \ge 1-\varepsilon$;
    \item $\Pr_{x\sim \unifcube}[h(x) = 1] \le \alpha\cdot n^{c+1}\poly\log(n/\delta)/\delta$.
\end{enumerate}
with probability at least $1-\delta$. 
Moreover, the algorithm 
\begin{enumerate}
    \item uses $m = \poly(n, 1/\varepsilon ,\log(1/\delta))$ i.i.d.\ samples from the target distribution $U_{L^\star}^+$,
    \item makes $O(\log(1/\wh \alpha)/\log (1/p))$ queries to the PAC learner for $C$ with $p$-negative-filtering, with filtered sample size $|S_p| = \poly(n,1/\varepsilon ,\log(1/\delta))$, confidence $\delta_\pac = \delta/n$, and runtime $T_\pac^p(n,|S_p| ,1/\delta_\pac)$
    \item at most $O(\log(1/\wh \alpha)/\log(1/p))$ queries to the approximate counting oracle for intersections of $k=O(\log(1/\wh \alpha)/\log(1/p))$ concepts in $C'$, with accuracy $\epsilon$, confidence $\delta'=\delta/n$, and runtime $T_{\cnt}^{\bigcap C'}(n, k, 1/\varepsilon ,1/\delta')$, 
    \item makes $O(\log(1/\wh \alpha)/\log(1/p))$ queries to the approximate uniform generation oracle for intersections of $k=O(\log(1/\wh \alpha)/\log(1/p))$ concepts in $C'$ with accuracy $\delta''=\poly(1/n,\delta)$ and runtime $T_{\gen}^{\bigcap C'}(n, k, 1/\delta'')$.
\end{enumerate}
The algorithm runs in time
\[
\poly(n, 1/\epsilon, \log(1/\delta),T_\pac^p(n,|S_p| ,1/\delta_\pac), 
T_{\cnt}^{\bigcap C'}(n, k, 1/\varepsilon ,1/\delta'),
T_{\gen}^{\bigcap C'}(n, k, 1/\delta''),
\]
\end{theorem}

{Let us comment on the query complexity and the number of intersections  $O(\log \wh \alpha / \log p)$ of our algorithm for a particular mass guess $\wh \alpha \in [2^{-n},1]$. Intuitively, both of these quantities appear due to the following two properties:
\begin{enumerate}
    \item The first property is the need of the PAC learning algorithm for negative examples: this is depicted by the inverse dependence on $\log (1/p)$ since as $p \to 0$, then learning with $p$-negative filtering corresponds to learning only from positive examples and hence the number of queries and intersections is small. That is, as $p \to 0$, the algorithm is stronger and $\log p \to -\infty$.

    \item The second property is the guessed mass of the true language on the satisfying assignments, which corresponds to $\wh \alpha$: this is reflected in the $\log(1/\wh \alpha)$ factor; as the guessed mass on satisfying assignments increases, the query and intersection complexity both decrease. Intuitively, if the guessed mass is small, i.e., $\wh \alpha = 2^{-\Theta(n)}$ with no ``robustness'' assumptions on the PAC learner (i.e., $p = 1)$, one needs roughly $\Theta(n)$ intersections.
\end{enumerate}
}

\begin{remark}
\cref{thm:den-tradeoff} interpolates between \cref{thm:PAC-learn-to-sample-OWF} and \cref{thm:densifier-ERM}. 
In \cref{thm:PAC-learn-to-sample-OWF}, the PAC learner only depends on the positive samples. One can think of $p = 0$, and the algorithm only requires approximate uniform generator for \emph{one} function in the concept class and {one sampling query}. 
In \cref{thm:densifier-ERM}, when there is no assumption on the PAC learner, one can think of $p = 1$, and thus the algorithm requires approximate uniform generator and approximate counter for intersections of $O(\log(1/\alpha))$ (where $\alpha = \Pr_{x \sim \unifcube}[L^\star(x) =1]$) functions in the concept class. 
\end{remark}

The algorithm works as follows:
During the execution of the algorithm, we keep predictions {$g^i = \bigcap_{j=0}^{i-1} h^j$} for the true language. Initially, we set $g^1=h^0 = 1$ (i.e., the language is the whole space). 
Given the values $\wh \alpha$ and $p$, we iterate for $T = \log(1/\wh \alpha)/\log(1/2p)$ rounds and at each round we perform the following steps, which will refine our prediction $h^i, i = 1,...,T$. 

In round 1, we will need to get a draw of $N_0$ labeled examples  which is the number of samples needed by our PAC learner $\cA_{\pac}^p$ {before filtering}. In particular, the positive subset will have size $N^+ \approx w N_0$, where $w$ is a parameter adapted {so that $\varepsilon _\pac/w = \varepsilon /n$}. Given numbers $N_0, N^+$, we draw $N^+$ i.i.d. samples from the true language. However, we do not have direct access to negative examples. Instead, the idea is to use our ``bad'' hypothesis $g^1 = h^0$ to generate simulated samples using the uniform generation oracle $\cA_{\gen}^{\bigcap C'}$ (using only one hypothesis in the intersection) and label all these examples negative (whose number is $N^- = N_0 - N^+)$. We then run the PAC learner to get a hypothesis $h^1$. In particular, we show in \Cref{lem:tv-sim-fil} that if we randomly draw few positive examples from $h^0$, {the TV distance between the simulated samples and filtered negative samples will be small, so that the performance of $\cA_\pac^p$ is guaranteed. We will update the prediction as $h^0 \cap h^1$ and repeat until the mass of $\bigcap_{j=0}^{i-1} h^j$ is comparable to the true mass. }

\begin{algorithm}[H]
\caption{Algorithm for \Cref{thm:den-tradeoff}}
\SetKwInput{Input}{Input}
\SetKwInput{Output}{Output}
\SetKwInput{Oracle}{Oracle Access}
\Input{Sample access to $U_{L^\star}^+$, parameters $\varepsilon ,\delta > 0$, a value $\wh  \alpha \in [2^{-n}, 1]$}

\Oracle{{$p$-negative-filtering PAC Learner $\cA_{\pac}^p$}, Sampling Oracle $ \cA_{\gen}^{\bigcap C'}$, Counting Oracle $ \cA_{\cnt}^{\bigcap C'}$}
$h^0 \gets \mathbf{1}$\;
\For{$i \gets 1$ \KwTo $\log(1/\wh \alpha)/\log(1/2p)$}{
    $\wh \alpha_i \gets \wh  \alpha(1+ \varepsilon )/ \cA_{\rm count}^{\bigcap C'}(i, h^0,\dots, h^{i-1},\varepsilon ,\delta/n)$\;
    $\eta \gets \max \{p, \wh \alpha_i n R(n, \delta/n)/\delta\}$\tcp*[r]{$R(n,\delta/n)= n^c \poly\log(n/\delta)$}
    $w \gets n\eta/(\eps+n\eta)$\;
    $\eps_{\pac} \gets \eps w/n$\;
    $N_0 \gets R(n, \delta/n)/\eps_{\pac}$, the sample size needed by $\cA_{\pac}^p$ before filtering to achieve $\eps_{\pac}$ accuracy\;
    $N^+ \sim \mathrm{Bin}(N_0, w)$\;
    $N^- \sim \mathrm{Bin}(N_0-N^+, p)$\;
    Draw $N^+$ \iid samples $x_1^+,\dots, x_{N^+}^+$ samples from $U^+_{L^\star}$\;
    Draw $N^-$ i.i.d.\ samples $x_1^-,\dots, x_{N^-}^-$ from $\cA^{\bigcap C'}_{\gen}(i, h^0,\dots, h^{i-1},\delta/(nN^-))$\;
    Run $h^i\gets \cA_{\pac}^p(\{(x_i^+, 1)\}_{i\in [N^+]}\cup\{(x_i^-, -1)\}_{i\in [N^-]}, \eps_{\pac},\delta/n)$ with $D = w U_{L^\star}^+ + (1-w) U_{g^i}^+$ where $g^{i} = \bigcap_{j=0}^{i-1} h^j$\;
    \If{$p \le \wh \alpha_i n R(n, \delta/n)/\delta$}{
        \Return{$h=\bigcap_{j=0}^i h^j$}\;
    }
}
\label{algo:den-tradeoffs}
\end{algorithm}

To see why we make progress, let us fix round $i$. Recall that at each round the PAC learner which is robust to $p$-negative filtering returns a hypothesis $h^i$ that achieves PAC learning guarantees with accuracy $\eps_{\pac}$ with respect to the base measure
    \[
    w U^+_{L^\star} + (1-w) U^+_{g^i}
    \]
    where $g^i =\bigcap_{j=0}^{i-1} h^j$. The PAC learning guarantee combined with \Cref{lem:tv-sim-fil} and the choice of $w$ imply that
    \begin{align*}
    \Pr_{x \sim U_{L^\star}^+} [h^i(x) \ne L^\star(x)] &\leq \frac{\varepsilon }{n}, \\
    \Pr_{x \sim U_{g^i}^+} [h^i(x) \ne L^\star(x)]  &\leq \eta.
\end{align*}
The first condition can be used to make progress on the breadth requirement since it controls the amount of true positive examples contained in $h^i$.  The second condition is for hallucination contraction: it controls the amount of positive examples in $g^i$, that are valid examples for the true language as well. Note that the above condition holds for a fixed $i$. One can then derive the desired breadth-hallucination guarantees of the main theorem by arguing about $g^i$, which is the intersection of all the previous PAC learners. For more details, we refer to the full proof of \Cref{thm:den-tradeoff}.

To make the above sketch formal, we first need to show that the simulated samples are close in TV distance to the filtered full samples.

\begin{lemma}
\label[lemma]{lem:tv-sim-fil}
For any $w, p,\delta \in (0, 1)$, $N_0 \ge 1$, $g:\{\pm1\}^n \to \{\pm 1\}$, let $N^+ \sim \bin(N_0, w)$, $N^- \sim \bin(N_0-N^+, p)$, and $x_1^+ ,\dots, x_{N^+}^+ \overset{\iid}{\sim} U_{L^\star}^+$,  $x_1^- ,\dots, x_{N^-}^- \overset{\iid}{\sim} \cD_{\rm gen}$ such that $\tv{U_{g}^+}{\cD_{\rm gen}} \le \delta_{\rm gen}$. And let $x_1,\dots, x_{N_0} \overset{\iid}{\sim}w U^+_{L^\star} + (1-w) U_g^{+}$. Then \begin{align*}
    \tv{\{(x_i^+, 1) \}_{i \in [N^+]} \cup \{(x_i^-, -1)\}_{i\in [N^-]}}{\cF_p(\{(x_i, L^\star(x_i)\}_{i\in [N_0]})} \le N_0 (1-w)(\alpha_g + p\delta_{\rm gen}),
\end{align*}
where $\alpha_g = \Pr_{x \sim U^+_g}[L^\star(x)=1]$. 
\end{lemma}
The interpretation of this lemma is the following: our algorithm at each round has a guess hypothesis $g^i$ which is much larger than the true language and is hence used in order to provide some artificial negative examples for $L^\star$ (these negative examples are essentially random positive examples from $g^i)$. The PAC learner of our \Cref{thm:den-tradeoff} is robust to $p$-negative filtering and so it works with data of the form $\cF_p(\{(x_i, L^\star(x_i)\}_{i\in [N_0]})$. However, we can only simulate up to some error these data using the positive examples from $L^\star$ and the positive examples from $g = g^i$ (which is our prediction at round $i$). Hence, the first term of the TV distance is the data that we feed to our PAC learner while the second argument is the ``ideal'' input to the PAC learner.  
\begin{proof}
We will first prove the TV distance bound for one sample. Let 
\begin{align*}
\cD_{\rm s}^{\rm gen} &= w  \cdot (U^+_{L^\star}, +) + (1-w) \cdot p  (\cD_{\rm gen},-) + (1-w)(1-p)  \cdot \bot, \\
    \cD_{\rm s} &= w  \cdot (U^+_{L^\star}, +) + (1-w) \cdot p  (U_g^+,-) + (1-w)(1-p)  \cdot \bot, \\
    \cD_{\rm f} &= w  \cdot (U_{L^\star}^+, +) + (1-w)  \alpha_g \cdot (U_{g \cap L^\star}^+, +) + (1-w)(1-\alpha_g) p\cdot (U_{g \backslash L^\star}^+,-) + (1-w)(1-\alpha_g)(1-p) \cdot \bot, 
\end{align*}
where $(\cD, \pm)$ denotes the law of $(x, \pm 1)$ where $x\sim \cD$, $U_{g \cap L^\star}^+$ denotes the uniform distribution over $\{x:g(x)=L^\star(x)=1\}$ and $U_{g \backslash L^\star}^+$ denotes the uniform distribution over $\{x:g(x)=1,L^\star(x)=0\}$. 
Then, since 
\begin{align*}
\{(x_i^+, 1) \}_{i \in [N^+]} \cup \{(x_i^-, -1)\}_{i\in [N^-]} &= \{\tilde x_i^{\rm s}\}_{i \in [N_0]} \backslash \{\bot\}, \quad \tilde x_i^{\rm s} \overset{\iid}{\sim} \cD_{\rm s}^{\rm gen},\\
\cF_p(\{(x_i, L^\star(x_i))\}_{i\in [N_0]}) &=\{\tilde x_i^{\rm f}\}_{i \in [N_0]} \backslash \{\bot\} , \quad \tilde x_i^{\rm f} \overset{\iid}{\sim} \cD_{\rm f},
\end{align*}
by the union bound and the data processing inequality, it suffices to show that $\tv{\cD_{\rm s}^{\rm gen}}{\cD_{\rm f}} \le (1-w)(\alpha_g + p\delta_{\rm gen})$.
Indeed, by direct computation, \begin{align*}
    \tv{\cD_{\rm s}}{\cD_{\rm f}} &\le \frac12 \left( (1-w)p\alpha_G +(1-w)\alpha_g + (1-w)(1-p)(1-(1-\alpha_g))  \right) = (1-w)\alpha_g, \\
    \tv{\cD_{\rm s}^{\rm gen}}{\cD_{\rm s}}  &\le (1-w)p\delta_{\rm gen}.
\end{align*}
\end{proof}

\begin{proof}[Proof of \cref{thm:den-tradeoff}]
Assume $\alpha \le \wh  \alpha \le (1+\varepsilon )\alpha$, where $\alpha = \Pr_{x \in \{\pm 1\}^n}[L^\star(x)=1]$. 
Let $g^i = \bigcap_{j=0}^{i-1} h^j$, and $\alpha_i = \Pr_{x \sim U_{g^i}^+}[L^\star(x)=1]$. 
Let $T = \log(1/\wh \alpha)/\log(1/2p)$.
In the for-loop, we will prove $\Pr_{x \sim U_{L^\star}^+}[g^i(x)=1] \ge 1- \varepsilon $ by induction \emph{for any index $i \in [T]$} (this guarantees the breadth requirement of our predicted language).
For $i=1$, this is true because $g_1 = h^0=\mathbf{1}$.
Assume this holds for $i \ge 1$. 
First, one can show that $\alpha_i \le \wh \alpha_i \le 2\alpha_i$, where we set
\[
\wh \alpha_i = \frac{\wh \alpha (1+\epsilon)}{\cA_{\cnt}^{\bigcap C}(i, h^0,...,h^{i-1}, \epsilon, \delta,n)}
\]
For simplicity let $F = \{x:L^\star(x)=1\}$, $G_i = \{x: g^i(x)=1\}$, and $|G'_i|/2^n$ be the output of $\cA_{\rm count}^{\bigcap C'}$ {with input $g^i$}.
Then 
by the induction hypothesis, we have $1 - \varepsilon \le \frac{|F\cap G_i|}{|F|} \le 1$. And 
with probability at least $1 - \delta/n$, $|G_i|/(1 +\varepsilon ) \le |G'_i| \le |G_i|(1+\varepsilon )$ {by the approximate counting oracle}.
Assume this happens, then \begin{align*}
    \alpha_i = \frac{|F \cap G_i|}{|G_i|} \le \frac{|F|}{|G_i|} \le \frac{\wh  \alpha (1+ \varepsilon ) 2^n}{|G'_i|} =\wh \alpha_i \le \frac{ (1+ \varepsilon )^3 |F|}{|G_i|} \le \frac{(1+ \varepsilon )^3|F \cap G_i|}{(1-\varepsilon )|G_i|}\le 2\alpha_i. 
\end{align*}
This proves that our predicted mass $\wh \alpha_i$ can be used to estimate the value $\alpha_i$, which measures the consistency of our prediction $g^i$.
By \Cref{lem:tv-sim-fil}, with $D$ being the mixture $w U_{L^\star}^+ + (1-w) U^+_{g^i}$
\begin{align*}
    \Pr[\Pr_{x\sim D}[h^i(x) \ne L^\star(x)] > \varepsilon_{\pac} ] \le \delta/n + N_0 (1-w) \alpha_i = \delta/n + \frac{R(n,\delta/n)}{\eta} \alpha_i \le \delta/n + \frac{R(n,\delta/n)}{\wh \alpha_i nR(n,\delta/n)/\delta} \alpha_i \le 2\delta/n.
\end{align*}
That is, with probability at least $1-2\delta/n$, the error of $h^i$ on $D$ is at most $\varepsilon $. Assume this happens, then since 
\begin{align*}
    \Pr_{x \sim D} [h^i(x) \ne L^\star(x)] = w \Pr_{x \sim U_{L^\star}^+} [h^i(x) \ne L^\star(x)] + (1-w) \Pr_{x \sim U_{g^i}^+} [h^i(x) \ne L^\star(x)] ,
\end{align*}
we have \begin{align*}
    \Pr_{x \sim U_{L^\star}^+} [h^i(x) \ne L^\star(x)] &\le \frac{\varepsilon _{\pac}}{w} = \frac{\varepsilon }{n}, \\
    \Pr_{x \sim U_{g^i}^+} [h^i(x) \ne L^\star(x)] &\le \frac{\varepsilon _{\pac}}{1-w} = \eta.
\end{align*}
The first condition is for the breadth requirement and the second condition is for the hallucination contraction. 

Then we can argue that the algorithm always terminates in the if-statement. 
In the case when $p > \wh \alpha_i nR(n,\delta/n)/\delta$, we get $\Pr_{x \sim U_{g^i}^+} [h^i(x) \ne L^\star(x)] \le \frac{\varepsilon _{\pac}}{1-w} = \eta=p$, we have \begin{align*}
    \Pr_{x\sim U_{g^i}^+}[h^i(x)=1] = \Pr_{x \sim U_{g^i}^+}[h^i(x)=1, L^\star(x)=0] + \Pr_{x \sim U_{g^i}^+}[h^i(x)=1, L^\star(x)=1] \le p+ \alpha_i \le 2p.
\end{align*}
That is $\Pr_{x \in \{\pm 1\}^n}[g^i(x)=1,h^i(x)=1] \le 2p\cdot \Pr_{x \in \{\pm 1\}^n}[g^i(x)=1]$. Suppose this is the case for all $i$ up to $\log(1/\wh \alpha)/\log(1/2p)$. Then in the last iteration it must hold that $\Pr_{x \in \{\pm 1\}^n}[g_{i}(x)=1] \le \wh \alpha \le (1 + \varepsilon )\alpha$. 
Thus, by the induction hypothesis $\wh \alpha_i \ge \alpha_{i} = \frac{|F \cap G_i|}{|G_i|} \ge \frac{(1- \varepsilon )|F|}{|G_i|} \ge \frac{1-\varepsilon }{1+\varepsilon } \ge \frac{p\delta}{n^{c+1} \poly\log(n/\delta)}=\frac{p\delta}{nR(n,\delta/n)}$. That is, the if-condition holds. 

Now we can prove the breadth and hallucination requirements. 
Note that $\log(1/\wh \alpha)/\log(1/2p) \le n$.
By the union bound, we get the breadth requirement \begin{align*}
    \Pr_{x \sim U_{L^\star}^+}[g^i(x) \ne 1] = \Pr_{x \sim U_{L^\star}^+}[\exists j \in [i], h^j(x) \ne L^\star(x)] \le \varepsilon. 
\end{align*}
This also proves the induction hypothesis. 
For the hallucination bounds, since in the last iteration, we have $\eta = \wh \alpha_i n R(n,\delta/n)/\delta$.
Therefore, \begin{align*}
\Pr_{x \in \{\pm 1\}^n}[g_{i+1}(x)=1] &\le (\eta + \alpha_i) \Pr_{x \in \{\pm 1\}^n}[g^i(x)=1] \\
&\lesssim n R(n,\delta/n)/\delta\cdot \alpha_i \Pr_{x \in \{\pm 1\}^n}[g^i(x)=1] \\
&= n R(n,\delta/n)/\delta \Pr_{x \in \{\pm 1\}^n}[g^i(x)=1, L^\star(x)=1] \\
&\le n^{c+1} \poly\log(n/\delta)/\delta \Pr_{x\in\{\pm 1\}^n} [L^\star(x)=1] .
\end{align*}

For the running time, first the number of iterations is at most $\log(1/\wh \alpha)/\log(1/2p) \le n$. 
Also, note that $\bin(N, p)$ can be generated in $\poly(X,\log(N),\log(1/p))$ time, where $X$ is the outcome, and $N_0 = \frac{R(n,\delta/n)}{\eta} + \frac{nR(n,\delta/n)}{\varepsilon} \le \frac{R(n,\delta/n)}{\wh \alpha_i nR(n,\delta/n)/\delta} + \frac{nR(n,\delta/n)}{\varepsilon} \lesssim 2^n + \frac{nR(n,\delta/n)}{\varepsilon}$,
so it suffices to show that in each iteration, $N^+$ and $N^-$ are at most $\poly(n,1/ \varepsilon ,\log(1/\delta))$ with probability at least $1-\delta / n$ .
Indeed, since $\E[N^+] = N_0w=nR(n,\delta/n)/\varepsilon =n^{c+1}\poly\log(n/\delta)/ \varepsilon $, 
by the Chernoff bound, 
$\Pr[N^+ \ge 2 N_0w ] \le \exp(-N_0w/3) \lesssim \delta/n$. 
And similarly, since $N^- \sim \bin(N_0, (1-w)p)$, with probability at least $1-\delta/n$,
$N^- \le 2 N_0 (1-w)p = 2pR(n,\delta/n)/\eta \le n^c \poly\log(n/\delta)$.
\end{proof}

Similarly as in \cref{sec:sq+guess}, we can combine \cref{thm:den-tradeoff} and \cref{thm:sq+guess} to get the PAC learning to sample result for this trade-off. 

\begin{theorem}
\label{thm:PAC-learn-sample-tradeoff}
Let $C$ be a concept class of $n$-variate Boolean concepts with induced class of languages $\cL$. 
Then there is an algorithm that PAC learns to sample $C$, given access to 
\begin{enumerate}
    \item an efficient PAC learning algorithm with $p$-negative-filtering that outputs a hypothesis in $C'$ and achieves error $R(n,\delta_\pac)/N_0$ as in \cref{thm:den-tradeoff},
    \item an efficient approximate uniform generation oracle for intersections of $O(n/\log(1/p))$ concepts in $C'$,
    \item an efficient approximate counting oracle for intersections of $O(n/\log(1/p))$ concepts in $C'$, and
    \item an efficient SQ learning algorithm for $C$.
\end{enumerate}   
That is, for any $L^\star \in \cL$, given $\varepsilon ,\delta > 0$ and $m = \poly(n, 1/\varepsilon , 1/\delta)$ i.i.d.\ samples $S$ from $U_{L^\star}^+$, the algorithm outputs a sampler {$\poly(n,1/\varepsilon ,1/\delta)$-time-computable} $G_S$, such that with probability at least $1-\delta$, \[\tv{{G_S(U)}}{U_{L^\star}^+} \le \varepsilon , \]
in $\poly(n, 1/\varepsilon ,1/\delta)$ time.
\end{theorem}

\subsection{ERM with sampling and counting oracles for \texorpdfstring{$C$}{C} does not suffice}
\label{sec:sig-hardness}
The reduction of \citet{de2014learning} in \cref{sec:resultDDS} uses an online learner, approximate uniform generation and counting oracles for $C$, and an SQ learner.
\Cref{thm:PAC-learn-sample-ERM} replaces the online learner by an ERM oracle, but it needs the approximate uniform generation and counting oracles for intersections of concepts in $C$.
In this section, we show that these oracles for intersections cannot be replaced by such oracles for $C$ in general.
Namely, we construct a concept class $C$ with an efficient ERM oracle, efficient exact uniform generation and counting algorithms, and an efficient SQ learning algorithm, such that no polynomial-time algorithm PAC learns to sample $C$.
The key tool is a unique signature scheme, as in the hardness results of \citet{de2014learning}.

As in \cref{sec:ows}, we use $\{0,1\}^n$ instead of $\{-1,1\}^n$ as the domain in this section.
We first recall signature schemes (see e.g., \cite{goldreich2004foundations,katz2020introduction}).
Let $\lambda$ be the security parameter, and let $r=r(\lambda)$, $\ell=\ell(\lambda)$, $k=k(\lambda)$ and $d=d(\lambda)$ be polynomials in $\lambda$ with $\ell\ge\lambda$ and $r,k,d\ge 1$, which are the lengths of seeds (used when generating the keys), messages, public keys and signatures, respectively.

\begin{definition}[Signature scheme]
\label{def:sig}
A signature scheme consists of three polynomial-time algorithms $\Gen$, $\Sign$ and $\Ver$ with the following properties.
\begin{enumerate}
    \item The key generation algorithm $\Gen$, using a uniformly random seed $s\in\{0,1\}^r$, outputs a public key $\pk(s)\in\{0,1\}^k$ and a secret key $\sk(s) \in \{0, 1\}^{\poly(\lambda)}$.
    \item The signing algorithm $\Sign$ is randomized. Given a message $m\in\{0,1\}^\ell$ and a secret key $\sk\in \{0, 1\}^{\poly(\lambda)}$, it outputs a signature $\sigma\in\{0,1\}^d$.
    \item The verification algorithm $\Ver$ is deterministic. Given a message $m\in\{0,1\}^\ell$, a string $p\in\{0,1\}^k$ and a signature $\sigma\in\{0,1\}^d$, it outputs a bit $\Ver(m,p,\sigma)\in\{0,1\}$.
    \item (Correctness) For every seed $s\in\{0,1\}^r$ and message $m\in\{0,1\}^\ell$, every possible output $\sigma$ of $\Sign$ given $m$ and $\sk(s)$ satisfies $\Ver(m,\pk(s),\sigma)=1$.
\end{enumerate}
\end{definition}
Note that $\Ver(m,p,\sigma)$ is defined for every $p\in\{0,1\}^k$, including strings $p$ with $p\neq\pk(s)$ for every seed $s$.
We use this below, since our construction defines a concept for every such string $p$.

We use the following security notion.
The adversary gets signatures on uniformly random messages and has to output a new message--signature pair. 

\begin{definition}[RMA security]
\label{def:rma}
A signature scheme is secure against random message attacks (RMA-secure) if the following holds for every polynomial $N=N(\lambda)$ and every polynomial-time algorithm $\cA$.
Let $s\sim\unif(\{0,1\}^r)$ and $m_1,\dots,m_N\overset{\iid}{\sim}\unif(\{0,1\}^\ell)$ be independent.
For $i\in[N]$, let $\sigma_i$ be the output of $\Sign$ given $m_i$ and $\sk(s)$, independently over $i$.
Let $(m,\sigma)=\cA(\pk(s),m_1,\dots,m_N,\sigma_1,\dots,\sigma_N)$.
Then
\[
\Pr\left[\Ver(m,\pk(s),\sigma)=1 \text{ and } m\notin\{m_1,\dots,m_N\}\right]\le\negl(\lambda),
\]
where the probability is over $s$, $m_1,\dots,m_N$ and the randomness of $\Sign$ and $\cA$, and $\negl(\lambda)\lesssim \lambda^{-c}$ for every $c>0$.
\end{definition}
RMA security is implied by the standard notion of existential unforgeability under adaptive chosen-message attacks \cite{goldreich2004foundations,katz2020introduction}, since an adversary that chooses the messages can choose them uniformly at random.

Moreover, we use unique signature schemes, introduced by \citet{goldwasser1992invariant}, in which every message has exactly one valid signature under every public key $\pk(s)$.
Uniqueness implies that there are exactly $2^\ell$ valid message--signature pairs under $\pk(s)$, which we use in \cref{lem:sig-sizes}.

\begin{definition}[Unique signature scheme]
\label{def:unique-sig}
A signature scheme is unique if for every seed $s\in\{0,1\}^r$ and message $m\in\{0,1\}^\ell$, there is exactly one $\sigma\in\{0,1\}^d$ with $\Ver(m,\pk(s),\sigma)=1$.
\end{definition}
By correctness, $\Sign$ then always outputs this $\sigma$ given $m$ and $\sk(s)$ (and thus is deterministic), and we denote it by $\Sign(m,\sk(s))$.
RMA-secure unique signature schemes are known under number-theoretic assumptions \cite{bellare1996exact,micali1999verifiable,boneh2001short,lysyanskaya2002unique,boneh2004short,dodis2005verifiable,hohenberger2010constructing}, and \citet{de2014learning} base their hardness results on such schemes.

We now state the main result of this section.
Items~1--3 provide the oracles of \cref{thm:PAC-learn-sample-ERM}, except that the approximate uniform generation and counting algorithms are for $C$ instead of intersections of concepts in $C$.

\begin{theorem}
\label{thm:sig-hardness}
Assume that there is an RMA-secure unique signature scheme.
Then there is a concept class $C$ of $n$-variate Boolean functions, where $n=k+\ell+d+r+1$ and $k$, $\ell$, $d$ and $r$ are the lengths of public keys, messages, signatures and seeds of the key generation algorithm $\Gen$, respectively, such that the following hold.
\begin{enumerate}
    \item (Sampling and counting are easy) $C$ has an efficient exact uniform generation algorithm and an efficient exact counting algorithm.
    \item (ERM is easy) $C$ has an efficient ERM oracle.
    \item (SQ learning is easy) $C$ has an efficient SQ learning algorithm.
    \item (Learning to sample is hard) No algorithm running in time polynomial in $n$ satisfies \cref{def:PAC} for $C$ with $\eps=1/4$ and $\delta=1/3$.
\end{enumerate}
\end{theorem}

Hence, \cref{thm:PAC-learn-sample-ERM} does not hold if its approximate uniform generation and counting oracles for intersections of concepts in $C$ are replaced by such oracles for $C$.
Similarly, the online learner in \cref{thm:dds} cannot be replaced by an ERM oracle.

We now construct the class $C$.
Fix an RMA-secure unique signature scheme, and recall that $n=k+\ell+d+r+1$.
We identify $\{0,1\}^n$ with $\{0,1\}^k\times\{0,1\}^\ell\times\{0,1\}^d\times\{0,1\}^{r+1}$ and write $x=(p,m,\sigma,t)$ for $x\in\{0,1\}^n$.
Recall that $\circ$ denotes concatenation.
For $p\in\{0,1\}^k$ and $s\in\{0,1\}^r$, let
\[
V_p=\left\{(p,m,\sigma,0^{r+1}) : m\in\{0,1\}^\ell,\ \sigma\in\{0,1\}^d,\ \Ver(m,p,\sigma)=1\right\}
\]
and
\[
W_{p,s}=
\begin{cases}
\left\{(p,m,\sigma,1\circ s) : m\in\{0,1\}^\ell,\ \sigma\in\{0,1\}^d,\ \Ver(m,p,\sigma)=0\right\} & \text{if } \pk(s)\neq p,\\
\emptyset & \text{if } \pk(s)=p.
\end{cases}
\]
Thus $V_p$ consists of the valid message--signature pairs under $p$ with last component $0^{r+1}$, and $W_{p,s}$ consists of the invalid ones with last component $1\circ s$. 
Let $L_{p,s}=V_p\cup W_{p,s}$, and $f_{p,s}$ be the concept with $f_{p,s}^{-1}(1)=L_{p,s}$. 
Then $C=\{f_{p,s}:p\in\{0,1\}^k,\ s\in\{0,1\}^r\}$.

We describe $f_{p,s}$ by the pair $(p,s)$, for every $p\in\{0,1\}^k$ and $s\in\{0,1\}^r$.
In particular, the ERM oracle outputs such a pair, and the uniform generation and counting algorithms get such a pair as input.
Given $(p,s)$ and $x\in\{0,1\}^n$, one can decide whether $x\in L_{p,s}$ with one call to $\Gen$ and one call to $\Ver$.

The proof of \cref{thm:sig-hardness} relies on two properties of $C$.
First, if $\pk(s)=p$, then $L_{p,s}=V_p$, and i.i.d.\ samples from $U^+_{V_p}$ are signatures on uniformly random messages.
So a sampler whose output law is close to $U^+_{V_p}$ outputs a signature on a new message with constant probability, which contradicts RMA security.
Second, if $\pk(s)\neq p$, then $L_{p,s}$ contains each pair $(m,\sigma)$ exactly once, so one can sample from $U^+_{L_{p,s}}$ without a secret key for $p$.
The ERM oracle exploits that the sets $W_{p,s}$ for different $s$ are disjoint.
Given i.i.d.\ samples from $U^+_{V_p}$, it outputs $f_{p,s'}$ for some $s'$ such that no sample lies in $W_{p,s'}$.
This concept labels every sample correctly, but $|L_{p,s'}|=2^{\ell+d}$ if $\pk(s')\neq p$, while $|V_p|=2^\ell$ (see \cref{lem:sig-sizes}).

Consistent with \cref{thm:PAC-learn-sample-ERM}, intersections of two concepts in $C$ are hard to sample from.
Indeed, for $s\neq s'$, we have $L_{p,s}\cap L_{p,s'}=V_p$, since $W_{p,s}$ and $W_{p,s'}$ are disjoint.
An approximate uniform generation algorithm for intersections, run on the descriptions $(\pk(s),0^r)$ and $(\pk(s),1^r)$, would then output valid message--signature pairs under $\pk(s)$ without any signed messages, which contradicts RMA security.

The following lemma describes the sets $L_{p,s}$ explicitly.
We use it for the uniform generation and counting algorithms, and for the reduction from forging signatures to learning to sample.

\begin{lemma}
\label[lemma]{lem:sig-sizes}
Let $p\in\{0,1\}^k$ and $s\in\{0,1\}^r$.
\begin{enumerate}
    \item If $\pk(s)=p$, then $L_{p,s}=V_p$, and the map $m\mapsto(p,m,\Sign(m,\sk(s)),0^{r+1})$ is a bijection from $\{0,1\}^\ell$ onto $L_{p,s}$.
    In particular, $|L_{p,s}|=2^\ell$.
    \item If $\pk(s)\neq p$, then the map
    \[
    (m,\sigma)\mapsto
    \begin{cases}
    (p,m,\sigma,0^{r+1}) & \text{if } \Ver(m,p,\sigma)=1,\\
    (p,m,\sigma,1\circ s) & \text{if } \Ver(m,p,\sigma)=0
    \end{cases}
    \]
    is a bijection from $\{0,1\}^\ell\times\{0,1\}^d$ onto $L_{p,s}$.
    In particular, $|L_{p,s}|=2^{\ell+d}$.
\end{enumerate}
\end{lemma}
\begin{proof}
We first prove Item~1.
Since $\pk(s)=p$, we have $W_{p,s}=\emptyset$ and hence $L_{p,s}=V_p$.
The map is a bijection onto $V_p$ by uniqueness (\cref{def:unique-sig}).

Next, we prove Item~2.
The map is injective since its image determines $(m,\sigma)$, and it is onto $L_{p,s}$ by the definitions of $V_p$ and $W_{p,s}$, since $\pk(s)\neq p$.
\end{proof}

\begin{proof}[Proof of \cref{thm:sig-hardness}]
We prove Items~1--4 of \cref{thm:sig-hardness} in order for the class $C$ constructed above.
Recall that $\Gen$, $\Sign$ and $\Ver$ run in time $\poly(\lambda)$, which is $\poly(n)$ since $\lambda\le\ell\le n$.

\paragraph{Sampling and counting.}
Given a description $(p,s)\in\{0,1\}^k\times\{0,1\}^r$, the uniform generation and counting algorithms first run $\Gen$ with seed $s$ to compute $\pk(s)$ and $\sk(s)$.
If $\pk(s)=p$, the uniform generation algorithm draws $m\sim\unif(\{0,1\}^\ell)$ and outputs $(p,m,\Sign(m,\sk(s)),0^{r+1})$.
Otherwise, it draws $(m,\sigma)\sim\unif(\{0,1\}^\ell\times\{0,1\}^d)$ and outputs $(p,m,\sigma,0^{r+1})$ if $\Ver(m,p,\sigma)=1$ and $(p,m,\sigma,1\circ s)$ otherwise.
By \cref{lem:sig-sizes}, the output has law $U^+_{L_{p,s}}$ in both cases.
So the algorithm satisfies \cref{def:apxsample} with $D=U^+_{L_{p,s}}$, and it runs in time $\poly(n)$.
Also by \cref{lem:sig-sizes}, the counting algorithm can output $\Pr_{x\sim\unif(\{0,1\}^n)}[f_{p,s}(x)=1]$ exactly.
This probability is $2^{\ell-n}$ if $\pk(s)=p$ and $2^{\ell+d-n}$ otherwise, so the algorithm satisfies \cref{def:apxcount}, and it runs in time $\poly(n)$ since it makes one call to $\Gen$.

\paragraph{Implementing the ERM oracle.}
Let $\{(x_i,y_i)\}_{i\in[N]}$ be the input of the ERM oracle (\cref{def:erm}), where $y_i\in\{\pm1\}$ and $x_i=(p_i,m_i,\sigma_i,t_i)$.
Every point of $L_{p,s}$ has first component $p$, and every point of $W_{p,s}$ has last component $1\circ s$.
We use these two facts to reduce the minimization over $C$ to at most $2N+1$ candidate descriptions.
Let $\cK=\{p_1,\dots,p_N\}$, and for $p\in\cK$, let
\[
\cT_p=\left\{s\in\{0,1\}^r: t_i=1\circ s \text{ for some } i\in[N] \text{ with } p_i=p\right\}.
\]
If $\cK\subsetneq\{0,1\}^k$, let $p_0$ be an arbitrary string in $\{0,1\}^k\setminus\cK$.
For $p\in\cK$ with $\cT_p\subsetneq\{0,1\}^r$, let $s_p$ be an arbitrary string in $\{0,1\}^r\setminus\cT_p$.
The ERM oracle computes the number of mistakes $\sum_{i\in[N]}\ind\{f_{p,s}(x_i)\neq y_i\}$ for each of the following candidate descriptions $(p,s)$, and outputs a candidate with the fewest mistakes.
\begin{enumerate}
    \item $(p_0,0^r)$, if $p_0$ is defined (when \(\cK \subsetneq\{0, 1\}^k\)).
    \item $(p,s)$ for $p\in\cK$ and $s\in\cT_p$.
    \item $(p,s_p)$ for $p\in\cK$, if $s_p$ is defined (when $\cT_p\subsetneq\{0,1\}^r$).
\end{enumerate}

We now show that the output minimizes the number of mistakes over $C$.
It suffices to show that for every description $(p,s)$, some candidate labels every $x_i$ as $f_{p,s}$ does.
If $p\notin\cK$, then $p_0$ is defined, and no $x_i$ lies in $L_{p,s}$ or in $L_{p_0,0^r}$.
So $f_{p,s}(x_i)=f_{p_0,0^r}(x_i)=-1$ for every $i\in[N]$.
If $p\in\cK$ and $s\notin\cT_p$, then $s_p$ is defined, and no $x_i$ lies in $W_{p,s}$ or in $W_{p,s_p}$.
Since $L_{p,s}=V_p\cup W_{p,s}$ and $L_{p,s_p}=V_p\cup W_{p,s_p}$, we get $f_{p,s}(x_i)=f_{p,s_p}(x_i)=1$ if $x_i\in V_p$ and $f_{p,s}(x_i)=f_{p,s_p}(x_i)=-1$ otherwise, for every $i\in[N]$.
In the remaining case $p\in\cK$ and $s\in\cT_p$, the description $(p,s)$ is itself a candidate.

Finally, we bound the running time.
Since $|\cK|\le N$ and $|\cT_p|\le N$, the lexicographically first strings in $\{0,1\}^k\setminus\cK$ and $\{0,1\}^r\setminus\cT_p$ can be found in time $\poly(n,N)$, and we take them as $p_0$ and $s_p$.
Since $\sum_{p\in\cK}|\cT_p|\le N$, there are at most $2N+1$ candidates.
For each candidate, counting its mistakes takes $N$ membership tests, each with one call to $\Gen$ and one call to $\Ver$.
So the ERM oracle runs in time $\poly(n,N)$.

\paragraph{Learning with statistical queries.}
Let $f^\star=f_{p^\star,s^\star}$ be the target concept and $D$ be a distribution over $\{0,1\}^n$.
The SQ learning algorithm first computes $p^\star$ bit by bit, and then either computes $s^\star$ or outputs a hypothesis that ignores $W_{p^\star,s^\star}$.
Let
\[
\rho=\Pr_{x\sim D}[f^\star(x)=1]
\]
and
\[
\rho_W=\Pr_{x\sim D}[x\in W_{p^\star,s^\star}].
\]
For $x=(p,m,\sigma,t)$ and $y\in\{\pm1\}$, let $q_0(x,y)=\ind\{y=1\}$ and $q'_0(x,y)=\ind\{y=1,\,t_1=1\}$.
For $j\in[k]$, let
\[
q_j(x,y)=\ind\{y=1,\,p_j=1\},
\]
and for $j\in[r]$, let
\[
q'_j(x,y)=\ind\{y=1,\,t_1=1,\,t_{j+1}=1\},
\]
where $p_j$ and $t_j$ denote the $j$-th bits of $p$ and $t$.
Every $x\in L_{p^\star,s^\star}$ has first component $p^\star$, and $x$ has $t_1=1$ if and only if $x\in W_{p^\star,s^\star}$, in which case $t=1\circ s^\star$.
Hence, we have
\begin{itemize}
    \item $\E_{x\sim D}[q_0(x,f^\star(x))]=\rho$, and $\E_{x\sim D}[q_j(x,f^\star(x))]=\rho\,\ind\{p^\star_j=1\}$ for $j\in[k]$;
    \item $\E_{x\sim D}[q'_0(x,f^\star(x))]=\rho_W$, and $\E_{x\sim D}[q'_j(x,f^\star(x))]=\rho_W\,\ind\{s^\star_j=1\}$ for $j\in[r]$.
\end{itemize}
The algorithm makes all queries with tolerance $\eps/10$ and proceeds as follows.
\begin{enumerate}
    \item Query $q_0$ and let $v_0$ be the answer. If $v_0\le\eps/2$, output the hypothesis $h\equiv-1$.
    \item For $j\in[k]$, query $q_j$, let $v_j$ be the answer, and let $\wh p_j=\ind\{v_j\ge\eps/5\}$.
    \item Query $q'_0$ and let $v'_0$ be the answer. If $v'_0\le\eps/2$, output the hypothesis $h$ with $h(x)=1$ if $x\in V_{\wh p}$ and $h(x)=-1$ otherwise.
    \item For $j\in[r]$, query $q'_j$, let $v'_j$ be the answer, and let $\wh s_j=\ind\{v'_j\ge\eps/5\}$. Output $f_{\wh p,\wh s}$.
\end{enumerate}

We now show that the output $h$ satisfies $\Pr_{x\sim D}[h(x)\neq f^\star(x)]\le\eps$.
If the algorithm stops in Step~1, then $\Pr_{x\sim D}[h(x)\neq f^\star(x)]=\rho\le v_0+\eps/10\le 3\eps/5$.
Otherwise, $\rho\ge v_0-\eps/10>2\eps/5$.
In Step~2, we then have $v_j\ge\rho-\eps/10>3\eps/10$ if $p^\star_j=1$ and $v_j\le\eps/10$ if $p^\star_j=0$, so $\wh p=p^\star$.
If the algorithm stops in Step~3, then $h$ and $f^\star$ differ exactly on $W_{p^\star,s^\star}$, since $L_{p^\star,s^\star}=V_{p^\star}\cup W_{p^\star,s^\star}$ and $V_{p^\star}\cap W_{p^\star,s^\star}=\emptyset$.
So $\Pr_{x\sim D}[h(x)\neq f^\star(x)]=\rho_W\le v'_0+\eps/10\le 3\eps/5$.
Otherwise, $\rho_W>2\eps/5$, and the argument of Step~2 with $\rho_W$ in place of $\rho$ gives $\wh s=s^\star$, so the output $f_{\wh p,\wh s}=f^\star$ makes no mistakes.
The algorithm is deterministic, makes $k+r+2$ queries with tolerance $\eps/10$, and each query and the output hypothesis can be evaluated in time $\poly(n)$.
So it is an efficient SQ learning algorithm (\cref{def:sq}).

\paragraph{Constructing a forger from a learner.}
Suppose for contradiction that an algorithm $\cG$ runs in time $\poly(n)$ and satisfies \cref{def:PAC} for $C$ with $\eps=1/4$ and $\delta=1/3$.
Let $N$ be the number of samples that $\cG$ uses, which is at most its running time $\poly(n)=\poly(\lambda)$, since $n=k+\ell+d+r+1$.
We construct an algorithm $\cA$ that breaks the RMA security of the signature scheme (\cref{def:rma}).
Given $\pk(s)$ and $(m_i,\sigma_i)_{i\in[N]}$, the algorithm $\cA$ proceeds as follows.
\begin{enumerate}
    \item Set $x_i=(\pk(s),m_i,\sigma_i,0^{r+1})$ for $i\in[N]$.
    \item Run $\cG$ on $S=(x_1,\dots,x_N)$ to obtain a sampler $G_S$, and draw $z\sim G_S(U)$.
    \item If $z=(\pk(s),m,\sigma,0^{r+1})$ for some $m\in\{0,1\}^\ell$ and $\sigma\in\{0,1\}^d$, output $(m,\sigma)$. Otherwise, output $(0^\ell,0^d)$.
\end{enumerate}
The algorithm $\cA$ runs in time $\poly(\lambda)$.

We now lower bound the probability that $\cA$ outputs a signature on a new message.
Fix the seed $s$ and let $p=\pk(s)$.
By \cref{lem:sig-sizes}, $L_{p,s}=V_p$ and the map $m\mapsto(p,m,\Sign(m,\sk(s)),0^{r+1})$ is a bijection onto $V_p$.
Since $m_1,\dots,m_N\overset{\iid}{\sim}\unif(\{0,1\}^\ell)$ and $\sigma_i=\Sign(m_i,\sk(s))$, the points $x_1,\dots,x_N$ are i.i.d.\ samples from $U^+_{L_{p,s}}=U^+_{V_p}$.
Since $f_{p,s}\in C$, with probability at least $2/3$, we have $\tv{G_S(U)}{U^+_{V_p}}\le 1/4$.
Let
\[
E=\left\{(p,m,\Sign(m,\sk(s)),0^{r+1}): m\in\{0,1\}^\ell\setminus\{m_1,\dots,m_N\}\right\}.
\]
By the same bijection, $E\subseteq V_p$ and $|E|\ge 2^\ell-N$, so $U^+_{V_p}(E)\ge 1-N/2^\ell$.
If $\tv{G_S(U)}{U^+_{V_p}}\le 1/4$, then conditioned on $x_1,\dots,x_N$ and the randomness of $\cG$, we have $z\in E$ with probability at least $3/4-N/2^\ell$.
If $z\in E$, then $\cA$ outputs $(m,\sigma)$ with $\Ver(m,p,\sigma)=1$ and $m\notin\{m_1,\dots,m_N\}$.
Hence, for every seed $s$,
\[
\Pr\left[\Ver(m,p,\sigma)=1 \text{ and } m\notin\{m_1,\dots,m_N\}\right]\ge\frac23\left(\frac34-\frac{N}{2^\ell}\right)=\frac12-\frac{2N}{3\cdot 2^\ell},
\]
where $(m,\sigma)$ is the output of $\cA$.
Averaging over $s\sim\unif(\{0,1\}^r)$, the probability in \cref{def:rma} is also at least $1/2-2N/(3\cdot 2^\ell)$.
Since $N$ is a polynomial in $\lambda$ and $\ell\ge\lambda$, this is at least $1/3$ for all sufficiently large $\lambda$, which contradicts the RMA security of the signature scheme.
\end{proof}

\section{Conclusion}
We believe that our work opens various exciting directions for future research. The central question is to understand for which classes of languages over $\{-1,1\}^n$, there is an algorithm that PAC learns to sample in sub-exponential time. There are many possible extensions of the model such as generation in continuous domains and beyond realizable settings. Finally, our work focused on oracle-efficient algorithm design; understanding how to efficiently implement these oracles for natural concept classes is an interesting question for future work.

\renewcommand*{\bibfont}{\small}

\printbibliography

\appendix
\addtocontents{toc}{\protect\setcounter{tocdepth}{3}}

\section{Connections to Other Objectives and Memorization}
\label[appendix]{sec:otherDefs}
In this section, we would like to draw some parallels of our \Cref{def:PAC} with the recent line of work on language generation in the limit, initiated by \citet{kleinberg2024language}. 
A natural question is how the formulation of learning in total variation (recall \Cref{def:PAC}) relates to the existing notions of consistency and breadth studied by \cite{kalavasis2025limits,charikar2024exploring} in the context of language generation in the limit. Let us first discuss some natural definitions that aim at capturing breadth and hallucinations in our PAC language generation setting.

\begin{definition}
[Consistent Generation]
\label{def:hal}
A sampler $G$ achieves $\epsilon$-\emph{hallucination rate} with respect to the true language $L^\star$ if:
     \[
     \mathrm{consistency}(G) := \Pr_{x \sim G(U)}[x \in L^\star] \geq 1-\epsilon\,.
     \]
\end{definition}
This means that the support of the generation $G$ contains at most an $\epsilon$ mass of words outside of the true language. 
 
\begin{definition}
[Breadth relative to Boolean Domain]
\label{def:breadthBoolean}
A sampler $G$ achieves \emph{$(1-\epsilon)$-breadth relative to the Boolean domain} with respect to the true language $L^\star$ if:
     \[
     \mathrm{breadth}(G) := \Pr_{x \sim U_{L^\star}^{+}}[x \in \mathrm{supp}(G)] \geq 1-\epsilon\,.
     \]
\end{definition}
In words, this means that the support of the generator $G$ contains at least an $(1-\epsilon)$ fraction of the support of the true language.

We are mainly interested in generators that are balanced in the sense that:
\begin{definition}
    [Balanced Generator]
{We say that a generator $G$ is balanced if its output distribution $G(U)$ is uniform on a subset $S$ of the Boolean hypercube.}
\end{definition}

Moreover, one can show that the combination of low hallucination rate and breadth implies total variation learning for balanced generators.

\begin{proposition}
\label{prop:TV-breadth-consistent}
    Assume that $G$ is a balanced sampler that achieves 
    $(1-\epsilon)$-breadth relative to the true language and $\epsilon'$-hallucination rate for the class $C$. Then $G$ PAC learns to sample $C$ in total variation distance $\max\{\epsilon,\epsilon'\}.$
\end{proposition}

\begin{proof}
Let $S$ be the support of the output distribution of $G$
and $L^\star$ be the target language. Then since the generator is balanced we can write its hallucination as
\[
\mathrm{1-consistency}(G) = \Pr_{x\sim G(U)}[x \notin L^\star] = \frac{|S \setminus L^\star|}{|S|} \leq \epsilon',
\]
and its lack of breadth as
\[
1-\mathrm{breadth}(G) = 
\Pr_{x\sim U_{L^\star}^+}[x \notin S] = \frac{|L^\star \setminus S|}{|L^\star|} \leq \epsilon.
\]
Now observe that
\[
\mathrm{TV}(G, U_{L^\star}^+) 
=
1- \sum_x \min\{G(x), U_{L^\star}^+(x)\}. 
\]
In the above sum, we only care about $x$ in the common support of $G$ and $L^\star$, otherwise $
\min\{G(x), U_{L^\star}^+(x)\}=0$. That is, when $x \in S \cap L^\star$, 
\[
\min\{G(x), U_{L^\star}^+(x)\} = \min\left\{\frac{1}{|S|}, \frac{1}{|L^\star|}\right\} . 
\]
Then the total variation can be expressed as
\[
\mathrm{TV}(G, U_{L^\star}^+)=
\max\left\{1-\frac{|S\cap L^\star|}{|S|}, 1-\frac{|S\cap L^\star|}{|L^\star|}\right\}. 
\]
But the first term in the maximum is exactly $1-\mathrm{consistency}(G) \le \varepsilon '$  while the second term is exactly $1-\mathrm{breadth}(G) \le \varepsilon $.
Therefore, the total variation is at most $\max\{\epsilon,\epsilon'\}.$
\end{proof}

Observe that it is trivial to get a sampler that (i) achieves breadth with respect to the Boolean domain by taking $\supp(G) = \{-1,1\}^n$ without any sample (extreme breadth but no guarantee regarding hallucinations) {or} (ii) achieves low hallucination rate by memorizing training examples (avoids hallucinations but no breadth). Finally, we can also get an algorithm that perfectly learns $c^*$ with $\Omega(2^n)$ samples by simply memorizing the observed samples. Our work aims to understand the setting where the learner using much fewer samples and sample-polynomial runtime.

\section{The Proof of \Cref{thm:stat}}
\label[appendix]{sec:proof-stat}
\begin{proof}
{Using \Cref{prop:TV-breadth-consistent}, we know that it suffices to obtain a sampler whose induced sampling distribution is the \emph{uniform} on some set of points (i.e., on some language) and satisfies two properties: it achieves $\epsilon$-hallucination rate (as in \Cref{def:hal}) and $(1-\epsilon)$-breadth relative to the Boolean domain language (as in \Cref{def:breadthBoolean}). We will now design such a sampler.}
Let $S$ be an i.i.d.\ sample drawn from $U_{c^\star}^{+}$. Let $H_{\mathrm{cons}}$ be the set of hypotheses in $C$ such that 
\[
H_{\mathrm{cons}} = \{h \in C : h(x) = 1~\forall x \in S\}.
\]
Our algorithm outputs the minimal consistent hypothesis, in the sense that
\[
g = \mathrm{argmin}_{h \in H_{\mathrm{cons}}} |\{x \in \{-1,1\}^n : h(x) = 1\}|,
\]
and let $G$ be the uniform distribution on $\{x : g(x) =1\}$. First, we will prove that if $|S| \geq \wt{\Omega}(\mathrm{VC}(C)/\epsilon)$, then $G$ satisfies \Cref{def:hal} and \Cref{def:breadthBoolean}.

\paragraph{Breadth is large.} Our goal is to show that
\[\mathrm{breadth}(G) := \Pr_{x \sim U_{L^\star}^{+}}[x \in \mathrm{supp}(G)] \geq 1-\epsilon\,.
   \] 
   It suffices to show that
\[
\Pr_{x \sim U_{L^\star}^{+}}[x \in \mathrm{supp}(G)] = \frac{|\mathrm{supp}(G) \cap P|}{|P|} \geq 1-\epsilon,
\]
where $P$ is the support of $U_{L^\star}^{+}$ (positive examples of $c^\star$). Hence, it suffices to show that
\[
\Pr_{x \sim U_{L^\star}^{+}}[g(x) \neq c^\star(x)] \leq \epsilon.
\]
By standard PAC learning, we know that if $n \geq \mathrm{VC}(C) \log(1/\epsilon)/\epsilon$, any consistent hypothesis satisfies this property (and hence $g \in H_{\mathrm{cons}}$ also does satisfy it).  Indeed, let $P = \{x:c^\star(x)=1\}$ and $Q_h := \{x:h(x)=1\}$.
For every hypothesis $h\in C$, define the set of true positives missed by $h$:
\[
R_h :=P  \setminus Q_h
=
\{x\in P  : h(x)=-1\}.
\]
Consider the class
\[
\mathcal R := \{R_h:h\in C\}
\]
and note that its VC dimension is at most $O(\mathrm{VC}(C))$. Now, if
\[
|S|
\ge
O\left(
\frac{\mathrm{VC}(C)\log(1/\epsilon)+\log(1/\delta)}{\epsilon}
\right),
\]
then with probability at least $1-\delta$ over the draw of
$S$, every ``bad'' set $R\in \mathcal R$ satisfying
\[
\Pr_{x\sim U^+_{L^\star}}[x\in R] \ge \epsilon
\]
contains at least one sample point from the training set $S$.

Let us condition on this event. Consider any consistent concept $h\in H_{\mathrm{cons}}$ with the sample $S$, i.e., $h(x)=1$ for every $x\in S$.
This implies that
\[
S \cap R_h=\emptyset.
\]
Hence, $R_h$ cannot have
$\Pr_{x \sim U^+_{L^\star}}[x \in R_h]$ at least $\epsilon$. Hence
\[
\Pr_{x\sim U^+_{L^\star}}[x\in R_h]
=
\Pr_{x\sim U^+_{L^\star}}[h(x)=-1]
\le \epsilon.
\]
Applying this to $h=g$, since $g\in H_{\mathrm{cons}}$, gives
\[
\Pr_{x\sim U^+_{L^\star}}[g(x)\neq c^\star(x)]
\le \epsilon.
\]
Thus
\[
\mathrm{breadth}(G) \ge 1-\epsilon.
\]

\paragraph{Hallucination Rate is small.} Our goal is to show that 
     \[
     \mathrm{consistency}(G) := \Pr_{x \sim G(U)}[x \in L^\star] \geq 1-\epsilon\,.
     \]
Using the same notation as before, it suffices to show that \[
\frac{|\supp(G) \cap P|}{|\supp(G)|} \geq 1-\epsilon,
\]
where recall that $P$ is the support of the true language.
By the choice of the ERM (which is chosen by minimizing the size of the positive examples) and the fact that $c^\star \in H_{\mathrm{cons}}$ by realizability, we know that $|\supp(G)| \leq |P|$. This means that
\[
\frac{|\supp(G) \cap P|}{|\supp(G)|} \geq 
\frac{|\supp(G) \cap P|}{|P|}.
\]
Hence, if we want to make this quantity at least $1-\epsilon$, it suffices to take $n \geq \mathrm{VC}(C)/\epsilon.$ 

\paragraph{Putting everything together.} To get a PAC learning to sample guarantee, we have to show that the distribution $G$ defined by our learning algorithm satisfies that with probability $99\%$ over the training set,
\[
\mathrm{TV}(G, U^+_{L^\star}) \leq \epsilon\,.
\]
To do that we have to use the fact that our sampler is uniform on the set $\{x : g(x) = 1\}$. Then
\[
\mathrm{TV}(G, U_{L^\star}^+) = \frac{1}{2}\sum_{x} \Big|\frac{1\{x \in \supp(G)\}}{|\supp(G)|} - \frac{1\{x \in P\}}{|P|} \Big| = 1 - \sum_{x \in P \cap \supp(G)} \min\{1/|P|, 1/|\supp(G)|\}\,.
\]
Since we know that 
\[
\frac{|\supp(G) \cap P|}{|\supp(G)|} \geq 
\frac{|\supp(G) \cap P|}{|P|} \geq 1-\epsilon,
\]
we get that under $\epsilon$-hallucination rate and $(1-\epsilon)$-breadth, the total variation between the two distributions is at most $\epsilon$.
Finally, one can obtain a sampler by using rejection sampling on the output hypothesis.
\end{proof}

\end{document}